\documentclass[acmsmall,screen]{acmart}
\AtBeginDocument{%
  \providecommand\BibTeX{{%
    \normalfont B\kern-0.5em{\scshape i\kern-0.25em b}\kern-0.8em\TeX}}}

\setcopyright{acmlicensed}
\acmYear{2022}
\acmDOI{10.1145/3511917}

\acmJournal{TOMM}
\acmVolume{1}
\acmNumber{1}
\acmArticle{1}
\acmMonth{1}
\acmPrice{15.00}

\usepackage{xspace}
\makeatletter
\DeclareRobustCommand\onedot{\futurelet\@let@token\@onedot}
\def\@onedot{\ifx\@let@token.\else.\null\fi\xspace}

\def\eg{\emph{e.g}\onedot} 
\def\ie{\emph{i.e}\onedot} 
 
\def\etc{\emph{etc}\onedot} 
 
\def\etal{\emph{et al}\onedot} \def\st{\emph{s.t}\onedot}
\makeatother

\begin{document}

% \title{Towards Improved Few-Shot Classification via Dense Classification and Attentive Pooling}
% \title{Enhancing Few-Shot Classification via Dense Classification and Attentive Pooling}
\title{Revisiting Local Descriptor for Improved Few-Shot Classification}

\author{Jun He}
\email{hyj.hfut.mail@gmail.com}
\author{Richang Hong}
\authornote{Corresponding author.}
\email{hongrc.hfut@gmail.com}
\author{Xueliang Liu}
\email{liuxueliang1982@gmail.com}
\affiliation{%
  \institution{Intelligent Interconnected Systems Laboratory of Anhui Province, Hefei University of Technology}
  \streetaddress{193 Tunxi Rd}
  \city{Hefei}
  \state{Anhui}
  \country{China}
  \postcode{230009}
}
\affiliation{%
  \institution{Institute of Artificial Intelligence, Hefei Comprehensive National Science Center}
  \streetaddress{5089 Wangjiang West Rd}
  \city{Hefei}
  \state{Anhui}
  \country{China}
  \postcode{230026}
}

\author{Mingliang Xu}
\email{iexumingliang@zzu.edu.cn}
\affiliation{%
  \institution{Zhengzhou University}
  \streetaddress{100 Science Ave}
  \city{Zhengzhou}
  \state{Henan}
  \country{China}
  \postcode{450001}
}
\author{Qianru Sun}
\email{qianrusun@smu.edu.sg}
\affiliation{%
  \institution{Singapore Management University}
  \streetaddress{80 Stamford Rd}
  \country{Singapore}
  \postcode{178902}
}
\renewcommand{\shortauthors}{He and Hong, et al.}

\begin{abstract}
Few-shot classification studies the problem of quickly adapting a deep learner to understanding novel classes based on few support images. In this context, recent research efforts have been aimed at designing more and more complex classifiers that measure similarities between query and support images, but left the importance of feature embeddings seldom explored. We show that the reliance on sophisticated classifiers is not necessary, and a simple classifier applied directly to improved feature embeddings can instead outperform most of the leading methods in the literature. To this end, we present a new method named \textbf{DCAP} for few-shot classification, in which we investigate how one can improve the quality of embeddings by leveraging \textbf{D}ense \textbf{C}lassification and \textbf{A}ttentive \textbf{P}ooling. Specifically, we propose to train a learner on base classes with abundant samples to solve dense classification problem first and then meta-train the learner on a bunch of randomly sampled few-shot tasks to adapt it to few-shot scenario or the test time scenario. During meta-training, we suggest to pool feature maps by applying attentive pooling instead of the widely used global average pooling (GAP) to prepare embeddings for few-shot classification. Attentive pooling learns to reweight local descriptors, explaining what the learner is looking for as evidence for decision making. Experiments on two benchmark datasets show the proposed method to be superior in multiple few-shot settings while being simpler and more explainable. 
% Code would be made public as soon as possible.
% Code is available at: \url{https://github.com/Ukeyboard/dcap/}.
Code would be made public at: \url{https://github.com/Ukeyboard/dcap/}.
\end{abstract}



\maketitle

\section{Introduction}
\label{sec:introduction}

Deep neural networks have achieved remarkable success in a variety of fields, such as object detection~\cite{lin2017feature,bochkovskiy2020yolov4}, segmentation~\cite{he2017mask,zhang2021self}, action recognition~\cite{shi2019skeleton,tan2021selective},
%person reidentification~\cite{yang2017person,yang2018person}, 
image captioning~\cite{bin2017adaptively,zhou2020more}, cross-modal retrieval~\cite{yang2021deconfounded,yang2022video} and especially image classification~\cite{szegedy2016rethinking,huang2017densely,tan2019efficientnet,meng2019learning}, since AlexNet~\cite{krizhevsky2012imagenet} won the ImageNet Large Scale Visual Recognition Challenge (ILSVRC) in 2012. However, the success hinges on the ability to apply gradient-based optimization routines to high-capacity models given hundreds if not thousands of training examples as supervision~\cite{santoro2016meta}. If the training data is scare, studies ~\cite{vinyals2016matching, finn2017model} have shown that deep learning models would fail to learn. In contrast, humans can learn fast even when taught by shown a few examples. To give deep learning models the generalization ability of humans, recently there has been an increasing interest in few-shot classification~\cite{ravi2017optimization,finn2017model,mishra2018snail,rusu2018metalearning,oreshkin2018tadam,liu2019learning,hou2019cross,jiang2020few,he2020memory,zhang2020uncertainty,simon2020adaptive,xu2021unsupervised}.

Few-shot classification aims to tackle the problem of learning a novel set of classes based on few support images. Generally, given a learner that is conceptually composed of an embedding network and a classifier, one straightforward approach towards this problem is transfer learning, \ie fine-tuning a pre-trained embedding network, as well as a newly initialized classifier, on the few support images so as to adapt them to the target recognition task, since transfer learning is proven effective in conventional supervised learning scenarios. Whereas, in few-shot setting, the naive learning procedure is at high risk of overfitting due to the lack of training examples. Data augmentation or regularization techniques~\cite{hariharan2017low,wang2018low} may alleviate overfitting by increasing data diversity or shrinking model searching space, yet they can not solve it. To tackle the challenge, Vinyals~\etal~\cite{vinyals2016matching} introduced an innovative learning paradigm called meta-learning, by following the idea of ``learning to learn''. Instead of directly optimizing the learner upon the few support images, meta-learning assumes the existence of a set of base classes with sufficient examples per class, divided into disjoint meta-training, meta-validation and meta-testing splits, and proposes to train the learner on a collection of few-shot tasks randomly sampled from the meta-training split (also referred as meta-training set), targeting at learning transferable knowledge from these similar tasks to facilitate the learning of novel classes when applied to a new task. Effective and elegant, meta-learning has become the de facto solution to few-shot classification.

While more and more meta-learning approaches are proposed, we notice that most of them concentrate on reformulating the classifier that is responsible for measuring the similarities between query and support images, but leave the feature embeddings that are used for classification seldom explored. 
For example, Vinyals~\etal~\cite{vinyals2016matching} proposed an attention-based classifier by comparing a query against all support images. Snell~\etal~\cite{snell2017prototypical} computed the squared Euclidean distances from the query to all category prototypes for classification. Sung~\etal~\cite{sung2018learning} and He~\etal~\cite{he2020memory} formulated the classifier as a relation network that learns to estimate the similarity of a query-support pair. Zhang~\etal~\cite{zhang2020uncertainty} in addition proposed an uncertainty-aware classifier, by taking observation noises in the query-support similarities into account. 
As for the feature embeddings, these methods typically apply global average pooling (GAP) over feature maps to simply prepare embeddings for the classifier that follows. 
We argue that the undue emphasis on the classifier in previous work overly undervalues the importance of feature embeddings.
Thus, in an orthogonal direction, we propose to tackle few-shot classification differently from the perspective of acquiring more discriminative and robust embeddings in this work. 
% We report that, when strong embeddings are given, the reliance on sophisticated classifier is not necessary and a simple one, such as nearest-centroid classifier, can work surprisingly well. 
% To be specific, we formulate meta-learning for few-shot classification as a two-stage training paradigm, namely pre-training and meta-finetuning. Whenever a set of novel classes with few support images and a set of base classes with sufficient examples are provided, our first step is to train the learner on the base classes to perform large-scale classification with standard cross-entropy loss. After that, we fine-tune the learner on randomly sampled few-shot tasks in the meta-training set to adapt it to few-shot scenario or the test time scenario. 
To be specific, we follow a two-stage training paradigm for model optimization and representation learning, namely pre-training and meta-finetuning. Whenever a set of novel classes with few support images and a set of base classes with sufficient examples are provided, our first step is to train the learner on the base classes to perform large-scale classification with standard cross-entropy loss. After that, we fine-tune the learner on randomly sampled few-shot tasks in the meta-training set to adapt it to the few-shot scenario or the test time scenario. 
% We demonstrate that the pre-training sets up a proper model initialization, from where one can derive a satisfactory baseline model by fine-tuning the pre-trained learner on just tens of thousands of few-shot training tasks. Meanwhile, we show that the meta-finetuning stage plays a crucially important role in the acquisition of fast generalization ability.
Without bells and whistles, we demonstrate that the two-stage training is able to achieve good results and, especially, the pre-training can bring more than $4\%$ improvement on average over models meta-trained from scratch, which indicates the effectiveness of good embeddings. 
% We also show that the meta-finetuning plays a crucially important role in the acquisition of fast generalization ability.

\begin{figure}[t!]
\begin{center}
    \includegraphics[width=0.8\linewidth]{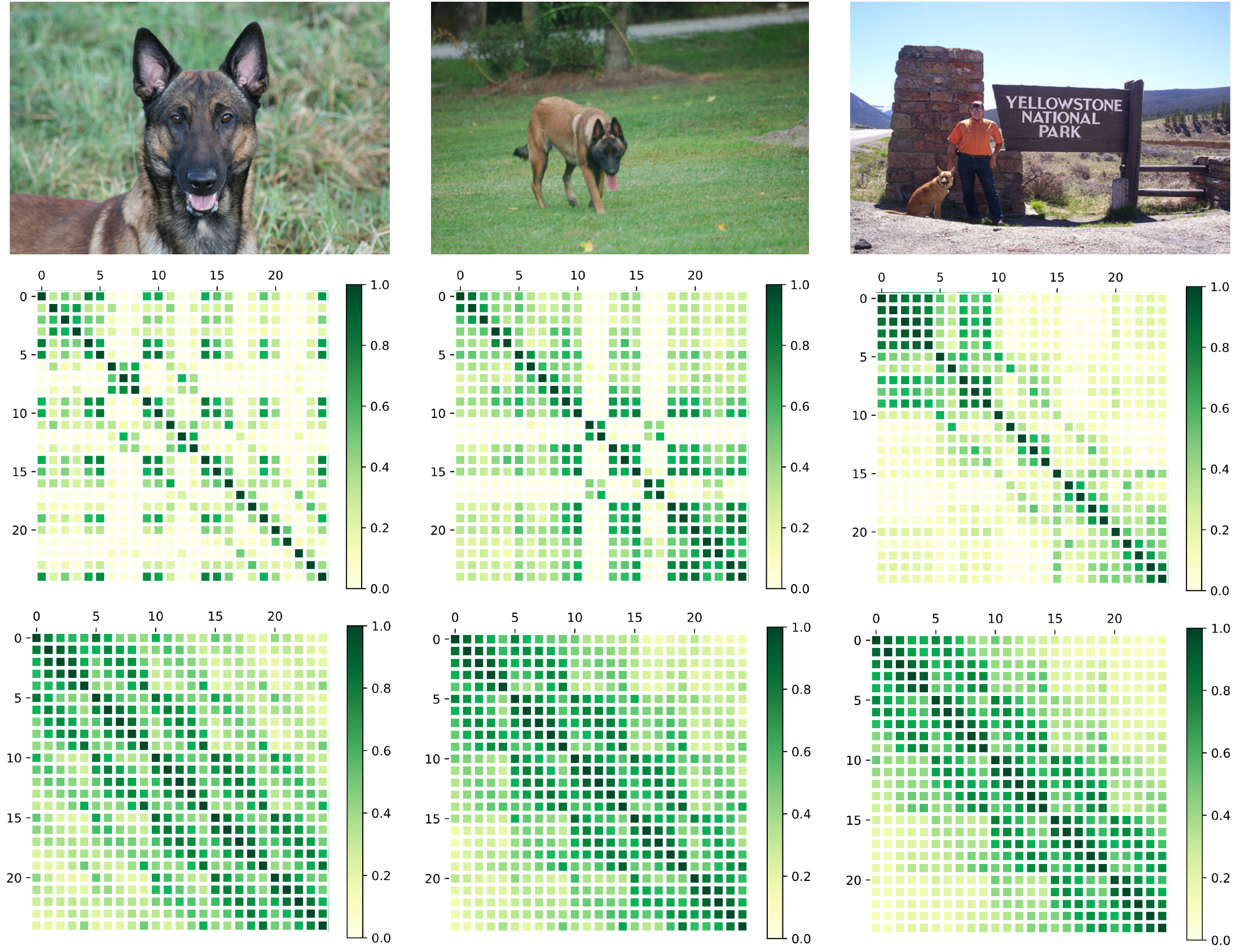}
\end{center}
% \vspace{-1em}
\caption{ \textbf{Cosine similarities between local descriptors.} From top to bottom: sample images, similarities between local descriptors in the feature map generated by a GAP-pretrained model, and similarities between local descriptors in the feature map generated by a DC-pretrained model. From left to right: salient target object in the middle (ideal example for training and testing), small target object in clear background, and small target object contaminated by distractors.}
\label{fig:distractor}
% \vspace{-1em}
\end{figure}

In addition to the two-stage training, we further propose to leverage dense classification and attentive pooling to improve the quality of feature embeddings. Particularly, we propose to perform dense classification over local descriptors in the pre-training stage. In meta-finetuning, we develop an attentive pooling strategy for summarizing local descriptors into a compact image-level embedding for few-shot classification. The proposition of dense classification is based on the observation that local descriptors within the same feature map generated by a GAP-pretrained model\footnote{A GAP-pretrained model refers to the model in which feature maps are pooled into feature embeddings by applying GAP before classification in the pre-training stage.} often exhibit severe semantic discrepancy\footnote{By semantic discrepancy, we mean two local descriptors have a small cosine similarity score (\eg 0.1) but they ought to have a big one (\eg 1.0).} (see Fig~\ref{fig:distractor}). The discrepancy is somewhat counterintuitive because convolutional layers are technically locally connected layers and each local descriptor in the final feature map thus generally has a large enough receptive field that probably covers the same semantic region of the input image. We show that dense classification in a DC-pretrained model\footnote{A DC-pretrained model refers to the model pre-trained with dense classification over local descriptors enabled in the pre-training stage.} can eliminate the discrepancy by enforcing semantic consistency among local descriptors. The idea of attentive pooling, on the other hand, is inspired by two insights. First, not all local image regions are equally representative to describe the target object in image. For example, as illustrated in Fig~\ref{fig:distractor}, the target object could only occupy a small proportion of the image, or even be contaminated by distractors. Second, the GAP strategy used in existing work does not deal with spatial information and forces to attach equal importance to each local descriptor, which not only causes information loss but also hinders the learner from attending to informative image regions, affecting the classifier negatively. By attentive pooling, we aim to select more informative local descriptors to build better image embeddings for the good of performance. We find that such a pooling strategy can consistently outperform its GAP counterparts.

We dub our method DCAP, \ie \textbf{D}ense \textbf{C}lassification pre-training plus meta-finetuning with \textbf{A}ttentive \textbf{P}ooling for few-shot classification. Our contributions are as follows: 
\begin{itemize}
    \item Firstly, we formulate meta-learning as a two-stage training paradigm, where we introduce a dense classification pre-training stage to reduce semantic discrepancy among local descriptors and devise an attentive pooling strategy in meta-finetuning to select more informative local descriptors for few-shot classification.
    \item Secondly, we justify the importance of pre-training in the acquisition of strong representation ability and validate the effectiveness of meta-finetuning in the acquisition of fast generalization ability. We show that \textit{(i)} pre-training can bring more than $4\%$ improvement on average over models meta-trained from scratch, but \textit{(ii)} a pre-trained model in inference mode performs no better than random guessing unless an extra meta-finetuning is carried out.
    \item Last but not least, we demonstrate that the reliance on sophisticated classifier in prior art is not necessary and set up a new baseline where a simple nearest-centroid classifier outperforms most of the leading methods by a large margin. On basis of empirical evaluation on \textit{mini}Imagenet~\cite{vinyals2016matching} and \textit{tiered}Imagenet~\cite{ren2018meta} datasets, we provide insights for best practices in implementation.
\end{itemize}

\section{Related Work}
\subsection{Meta-learning for Few-shot Classification}
Meta-learning provides a paradigm where a learner gains experience over plenty of related but distinguished tasks and thereafter uses the experience to efficiently solve similar but previously unseen tasks~\cite{xu2021exploring}. This `learning-to-learn' is recognized to show many benefits, including but not limited to data and compute efficiency~\cite{hospedales2020survey}. To learn from few examples, few-shot classification thus has widely adopted this very paradigm and achieved promising results. Approaches in this line can be broadly grouped into two: optimization-based~\cite{ravi2017optimization,finn2017model,li2017metasgd,nichol2018reptile,rusu2018metalearning,lee2019meta,baik2020alfa} and metric-based~\cite{vinyals2016matching,snell2017prototypical,oreshkin2018tadam,li2020boosting,he2020memory,jiang2020few}. 
In this part, we briefly review the two branches and cite some known work in the literature. For a complete list of recent work on meta-learning, we refer readers to~\cite{bendre2020survey, hospedales2020survey}.

Optimization-based approaches intend to train a meta-learner either to learn a general optimization algorithm as an alternative of the SGD optimizer for updating the target learner during training, or to learn a good initialization of the learner that allows for quick adaptation to a new task in a small number of model updating with a small amount of training data. 
Representative work in this branch include Meta-learner LSTM~\cite{ravi2017optimization}, MAML~\cite{finn2017model} and the variants of MAML, such as Reptile~\cite{nichol2018reptile}, LEO~\cite{rusu2018metalearning}, MetOptNet~\cite{lee2019meta} and ALFA~\cite{baik2020alfa}, to name a few.
% In optimization-based approaches, tremendous efforts have been made to advancing the learning paradigm, but with little attention being paid to feature embeddings.
In Meta-learner LSTM, the authors described an LSTM-based model that is trained to discover both a task-common initialization for the learner's parameters, as well as an update rule for updating the learner's parameters.
MAML proposed a tiered meta-learning paradigm in which a good initialization of the learner's parameters is slowly extracted in its outer-loop training and a fast adaptation within several gradient steps is explicitly enforced in the inner-loop fine-tuning.
LEO recognized the difficulties of directly operating on high-dimensional parameters in few-shot setting. Consequently, it learned a low-dimensional latent space in which optimization-based meta-learning is performed and from which the high-dimensional parameters can be generated. 
While optimization-based approaches have advanced few-shot classification a lot, the efforts they have made are largely made to advance the meta-learning algorithm~\cite{li2017metasgd,nichol2018reptile,baik2020alfa}, with little attention being paid to feature embeddings.
Lately, Raghu~\etal~\cite{raghu2019rapid} investigated the effectiveness of MAML and suggested that the performance of optimization-based meta-learning is mainly determined by the quality of the learned embeddings. 
This finding is very coherent with our emphasis on the importance of high-quality feature embeddings.

Metric-based approaches normally train the learner to learn an easily transferable embedding space, where ideally datapoints in the same class keep close to each other and datapoints from different classes stay far apart. 
The core idea is related to nearest neighbor algorithms and deep metric learning~\cite{liu2020deep,yang2018person}. 
In general, metric-based approaches would transform input data into lower dimensional embeddings first and then use the embeddings to design proper classifier for clustering or measuring their similarities. 
For example,
Vinyals~\etal~\cite{vinyals2016matching} presented Matching Networks which employs an attention mechanism to derive a weighted k-NN classifier. 
Snell~\etal~\cite{snell2017prototypical} took the mean vector of support images belonging to the same class as prototype and introduced a nearest-centroid classifier, \ie Prototypical Networks.
TADAM~\cite{oreshkin2018tadam} introduced metric-scaling, task conditioning and auxilary co-training to boost the performance of Prototypical Networks. 
Li~\etal~\cite{li2020boosting} further introduced an adaptive margin loss to help separate samples from different classes. 
To overcome limitations of using pre-defined distance metrics (\eg cosine distance~\cite{vinyals2016matching} and Euclidean distance~\cite{snell2017prototypical}) for classification, Sung~\etal~\cite{sung2018learning} and He~\etal~\cite{he2020memory} also formulated the distance metric as a learnable network in order to learn a generic metric that better suits few-shot setting. 
% Zhang~\etal~\cite{zhang2020uncertainty} proposed an uncertainty-aware classification framework that models uncertainty and performs uncertainty-aware optimization, by taking observation noises in similarity measurement into account.
In this work, we employ a simple nearest-centroid classifier that is similar to the one in Prototypical Networks for few-shot classification. But unlike Prototypical Networks who are famed for the proposed classifier, we focus on producing more powerful embeddings to support the classifier, by explicitly proposing to leverage dense classification and attentive pooling to improve the quality of feature embeddings. 

\subsection{Towards Improving Feature Embeddings}
% There is another group of metric-based approaches that devote to tackle few-shot classification from the perspective of improving feature embeddings.
% Except for the aforementioned classifier-centered metric-based approaches, there is another group of approaches that devote to improve feature embeddings.
% As a complement to the aforementioned classifier-centered approaches, there is another group of meta-learning approaches that devote to improving feature embeddings.
There is a family of meta-learning approaches that devote to improving feature embeddings.
Satorras and Estrach~\cite{satorras2018gnn} and Kim~\etal~\cite{kim2019edge} explored graph neural networks in the context of few-shot classification and presented graphical models that enhance embeddings by borrowing information from neighbors in a graph.
Similarly, ~\cite{li2019memory,he2020memory,rodriguez2020embedding} presented graph-like transductive algorithms that take all unlabeled query images into the embedding propagation procedure so that information in the unlabeled query images can be leveraged for embedding enhancement.
~\cite{hariharan2017low, wang2018low} viewed the problem differently. They presented generative algorithms to generate new training examples, and proposed to learn better embeddings from more training examples with larger intra-class diversity.
In~\cite{su2020does}, Su~\etal studied the effectiveness of utilizing self-supervised learning (SSL) techniques in few-shot setting. 
SSL utilizes the structure information contained already in an image to facilitate the learning of representations, having been well studied in traditional unsupervised learning with large unlabeled datasets~\cite{doersch2015unsupervised,larsson2016learning,gidaris2018unsupervised}.
% The arising star in unsupervised representation learning, \ie contrastive learning~\cite{he2020momentum,chen2020simple,chen2020exploring}, is also utilized in recent few-shot classification methods~~\cite{tian2020rethinking,li2020few,majumder2021revisiting}.
Pre-training on large-scale dataset is able to yield embeddings of high quality. Dhillon~\etal~\cite{dhillon2019baseline} and Chen~\etal~\cite{chen2020new} both showed that fine-tuning a well pre-trained embedding network can produce results rivaling far more complex algorithms in the low-data regime. Tian~\etal~\cite{tian2020rethinking} also demonstrated that learning supervised or self-supervised representation on a large-scale dataset, followed by learning a linear classifier on top of this representation, could outperform leading few-shot approaches.

\subsection{Local Descriptors in Few-shot Classification}
In existing few-shot classification approaches, feature maps are usually pooled or flattened into embedding vectors before classification. 
In~\cite{li2019revisiting}, Li~\etal proposed to leverage local descriptors for better classification. They built an image-to-class classifier in which each class is represented by all local descriptors from support images belonging to that class, and computed the image-to-class similarity score by summing the similarities between each local descriptor in a query image and its k-nearest neighbors in the local descriptor space of that class.
Wu~\etal~\cite{wu2019parn} introduced a deformable embedding network to embed an image as a set of local descriptors. They in addition proposed to consider spatial relationship of semantic objects in two images when computing similarity scores for classification.
Likewise, given an image represented as a set of local descriptors, \cite{hou2019cross} and \cite{jiang2020few} proposed to adaptively match two images to highlight relevant regions so as to build more discriminative embeddings.
We donot follow the complex image-to-image adaptation but instead present a simple attentive pooling to highlight informative regions. 
The assumption is that, relevant regions in two images would stand out spontaneously if they are the key to correct classification.
% Our nearest-centroid classifier is much simpler than the image-to-class classifier proposed in~\cite{li2019revisiting}.

\subsection{Attention in Few-shot Classification}
An early adoption of attention in few-shot classification is Matching Networks~\cite{vinyals2016matching}, where Vinyals~\etal proposed to attentively propagate label information from support images to construct the logits of a query image for classification.
Laterly, Wu~\etal~\cite{wu2019parn} proposed a dual correlation attention mechanism to overcome sensitivity to spatial locations of semantic objects in two images when comparing them for similarity measurement in the relation network.
Hou~\etal~\cite{hou2019cross} presented a cross attention mechanism to highlight relevant regions in two images to enhance feature discriminability.
Jiang~\etal~\cite{jiang2020few} developed an attention-based meta-reweighting strategy to adapt the embeddings of query images by incorporating the embeddings of support images.
The image-to-class classifier proposed in~\cite{li2019revisiting} can also be explained as an application of attention mechanism, especially in its finding k-nearest neighbors of a query local descriptor.
Other related work include graph-based methods~\cite{satorras2018gnn,liu2019learning, kim2019edge, li2019memory, he2020memory, rodriguez2020embedding} and Transformer-based method~\cite{doersch2020crosstransformers}.
Graph-based methods are intrinsically related because the message passing mechanism in graph is usually interpreted as an attention-based information propagation action.
As for Transformer~\cite{vaswani2017attention}, it is a born attention-based framework.

\section{Methodology}
We consider a learner that adopts a nearest-centroid classifier and propose to train the learner by following the two-stage training paradigm described in \S\ref{sec:introduction}, \ie dense classification pre-training and meta-finetuning with attentive pooling. 
We find that the two-stage training ensures a set of high-quality embeddings, upon which even the simple classifier can achieve results that are competitive or better than most of the leading few-shot learning approaches.
In this section, we start by establishing preliminaries about problem definition and meta-learning in \S\ref{sec:fsc_metalearning}; then we introduce the nearest-centroid classifier employed by our method in \S\ref{sec:protonetwork}; finally, we detail the two-stage training paradigm in \S\ref{sec:dense_classification} and \S\ref{sec:metafinetuning}. 

\subsection{Few-shot Classification with Meta-learning}
\label{sec:fsc_metalearning}
A formal definition of few-shot classification considers the well-organized $N$-way $K$-shot few-shot setting, where the target task $\mathcal{T} = \{\mathcal{D}^{train},\mathcal{D}^{test}\}$ contains $N_s = N \times K$ labelled support images $\mathcal{D}^{train} = \{(\mathbf{x}_i^s, y_i^s)\}^{N_s}_{i=1}$ with $y_i^s \in \mathcal{C}_{novel}$ and $N_q$ unlabeled query images $\mathcal{D}^{test} = \{\mathbf{x}_i^q\}^{N_q}_{i=1}$ from the same label space. $\mathcal{C}_{novel} = [c_1, c_2, ..., c_N]$ is the set of novel classes to learn in the task. Assuming a learner $\mathcal{A}(\cdot ; \Theta)$ that is conceptually composed of two parts: an embedding network $f_{\theta}: \mathcal{X} \rightarrow \mathbb{R}^{d \times h \times w}$ and a classifier $g_{\varphi}: \mathbb{R}^d \rightarrow \mathbb{R}^N$, few-shot classification aims to adapt the learner to correctly label each query image in $\mathcal{D}^{test}$ based on the supervision provided in $\mathcal{D}^{train}$ with the following objective
\begin{equation}
    \hat{\Theta} = \arg \min_{\Theta} \mathcal{L}(\mathcal{D}^{test} \mid \mathcal{D}^{train}; \Theta) + \mathcal{R}(\Theta),
\end{equation}
where $\Theta=(\theta, \varphi)$ is the learner's parameters, $\mathcal{L}$ is the loss function and $\mathcal{R}$ is the regularization term, \eg weight decay. 

Optimizing the learner directly on the few support images in $\mathcal{D}^{train}$ is at high risk of overfitting.
Meta-learning presents an efficient way for model optimization in such an extremely low-data regime. 
In meta-learning, we would resort to an extra dataset $\mathcal{D}^{base}$ that has a disjoint label space $\mathcal{C}_{base}$ (\ie  $\mathcal{C}_{base} \cap \mathcal{C}_{novel} = \emptyset$) with sufficient samples per class, and train the learner on a collection of few-shot tasks randomly sampled from $\mathcal{D}^{base}$. 
Conventionally, the dataset $\mathcal{D}^{base}$ is divided into disjoint meta-training, meta-validation and meta-testing splits. 
From the meta-training split, we are allowed to sample a set of training tasks $\mathcal{D}_{meta}^{train} = \{(\mathcal{D}_i^{train}, \mathcal{D}_i^{test})\}_{i=1}^{I}$, where each task $\mathcal{T}_i = (\mathcal{D}_i^{train}, \mathcal{D}_i^{test})$ instantiates an $N$-way $K$-shot few-shot classification problem that mimics the target task $\mathcal{T}$.
The learner is then trained to minimize the average test error on $\mathcal{D}_{meta}^{train}$, \ie
% \begin{equation}
%     \resizebox{.91\linewidth}{!}{$
%     {\Theta}^{\star} = \arg \min_{\Theta} \mathbb{E}_{\mathcal{T}_i\in\mathcal{D}_{meta}^{train}}[\mathcal{L}(\mathcal{D}_i^{test} \mid \mathcal{D}_i^{train}; \Theta)] + \mathcal{R}(\Theta).
%     $}
% \end{equation}
\begin{equation}
    {\Theta}^{\star} = \arg \min_{\Theta} \mathbb{E}_{\mathcal{T}_i\in\mathcal{D}_{meta}^{train}}[\mathcal{L}(\mathcal{D}_i^{test} \mid \mathcal{D}_i^{train}; \Theta)] + \mathcal{R}(\Theta).
\end{equation}
We also randomly sample a family of few-shot tasks $\mathcal{D}_{meta}^{val}$/$\mathcal{D}_{meta}^{test}$ from the meta-validation/meta-testing split.
$\mathcal{D}_{meta}^{val}$ is used to monitor the learning process for model selection.
And, the average test error on $\mathcal{D}_{meta}^{test}$ is taken as an indicator of how the learner generalizes to new tasks. 
By learning from similar tasks, meta-learning encourages the learner to learn transferable knowledge that facilitates the learning of novel classes when applied to the target task $\mathcal{T}$.

% \subsection{Prototypical Networks}
\subsection{The Nearest-centroid Classifier}
\label{sec:protonetwork}

In Prototypical Networks~\cite{snell2017prototypical}, Snell~\etal proposed a classifier that labels a query image based on its distance to each class centroid. Formally, given the task $\mathcal{T}$, the classifier is defined as follows
\begin{equation}
    y_i^q = \arg \max_{t} \frac{e^{s(\mathbf{f}_i^q, \mathbf{c}_t)/\tau}}{\sum_{j=1}^N e^{s(\mathbf{f}_i^q, \mathbf{c}_j)/\tau}}, ~~t=1,2,\dotsc,N
    \label{eq:protonet}
\end{equation}
where $\mathbf{f}_i^q \in \mathbb{R}^d$ denotes the feature embedding of image $\mathbf{x}_i^q \in \mathcal{D}^{test}$, $[\mathbf{c}_1,\mathbf{c}_2, \dotsc, \mathbf{c}_N]$ are the class centroids, $s(\cdot, \cdot)$ is a similarity function and $\tau$ is the temperature hyperparameter. To clarify, $\mathbf{f}_i^q$ is given by applying GAP over feature map $f_{\theta}(\mathbf{x}_i^q) \in \mathbb{R}^{d \times h \times w}$, \ie $\mathbf{f}_i^q = GAP(f_{\theta}(\mathbf{x}_i^q))$. The similarity function $s(\cdot, \cdot)$ could be cosine similarity or negative Euclidean distance. As for class centroid $\mathbf{c}_j$, it takes the average of support images in the $j$-th class (\ie $\mathcal{S}_j$) as defined by
\begin{equation}
    \mathbf{c}_j = \frac{1}{\vert\mathcal{S}_j\vert} \sum_{\mathbf{x}^s\in\mathcal{S}_j} GAP(f_{\theta}(\mathbf{x}^s)).
    \label{eq:centroid}
\end{equation}
% In this work, we adopt a similar nearest-centroid classifier that is as simple as Prototypical Networks. We show that a simple classifier can work surprisingly well if strong embeddings are given.
Simple but effective, as can be seen in \S\ref{sec:experiment}, Prototypical Networks provide a strong baseline for few-shot classification.

In this work, we focus on showing the importance of feature embeddings but not devote to designing as complex a classifier as the many other prior work did, so we adopt a similar nearest-centroid classifier here for few-shot classification. On top of Prototypical Networks, we make the following changes. 
First, we particularly propose to replace the GAP pooling with an attentive pooling strategy for summarizing local descriptors into image-level embeddings, \ie $\mathbf{f}_i^q = AttPool(f_{\theta}(\mathbf{x}_i^q))$.
Second, in order to reduce the labor required in hyperparameter tuning, we reformulate (\ref{eq:protonet}) as follows to eradicate hyperparameter $\tau$,
\begin{equation}
    y_i^q = \arg \max_{t} \frac{e^{\langle\mathbf{f}_i^q , \mathbf{c}_t/{\parallel \mathbf{c}_t \parallel}_2\rangle}}{\sum_{j=1}^N e^{\langle\mathbf{f}_i^q , \mathbf{c}_j/{\parallel \mathbf{c}_j \parallel}_2 \rangle}}, ~~t=1,2,\dotsc,N
    \label{eq:protonet_v2}
\end{equation}
where the dot product $\langle\mathbf{f}_i^q , \mathbf{c}_j/{\parallel \mathbf{c}_j \parallel}_2\rangle$ can be viewed as the projection of vector $\mathbf{f}_i^q$ on the $\mathbf{c}_j/{\parallel \mathbf{c}_j \parallel}_2$ axis.

\subsection{Dense Classification Pre-training}
\label{sec:dense_classification}
The pre-training refers to training the embedding network and a global classifier on the meta-training split for large-scale classification with standard cross-entropy loss. Let $\mathcal{D}^{train}_{base}$ be the meta-training split, prior work~\cite{dhillon2019baseline,sun2019meta,chen2020new} train with the following objective in this stage
% \begin{equation}
%     \resizebox{.91\linewidth}{!}{$
%     \displaystyle
%     \hat{\Phi} = \arg \min_{\Phi} \sum_{(\mathbf{x}, y_b) \in \mathcal{D}^{train}_{base}} \mathcal{L}^{ce}(\mathbf{W}^T GAP(f_{\theta}(\mathbf{x})) + \mathbf{b}, y_b) + \mathcal{R}(\Phi),
%     $}
%     \label{eq:gap-pretraining}
% \end{equation}
\begin{equation}
    \displaystyle
    {\Theta}^{\star} = \arg \min_{\Theta} \sum_{(\mathbf{x}, y_b) \in \mathcal{D}^{train}_{base}} \mathcal{L}^{ce}(\mathbf{W}^T GAP(f_{\theta}(\mathbf{x})) + \mathbf{b}, y_b) + \mathcal{R}(\Theta),
    \label{eq:gap-pretraining}
\end{equation}
where $\Theta = (\theta, \mathbf{W}, \mathbf{b})$ denotes learnable parameters including $\theta$ of the embedding network and $(\mathbf{W}, \mathbf{b})$ of the global classifier. Differently, we propose to remove the GAP operator and perform dense classification over local descriptors, \ie
% \begin{equation}
%     \resizebox{.91\linewidth}{!}{$
%     \displaystyle
%     \hat{\Phi} = \arg \min_{\Phi} \sum_{(\mathbf{x}, y_b) \in \mathcal{D}^{train}_{base}}\sum_{j=1}^r \mathcal{L}^{ce}(\mathbf{W}^T f^j_{\theta}(\mathbf{x}) + \mathbf{b}, y_b) + \mathcal{R}(\Phi),
%     $}
%     \label{eq:dc-pretraining}
% \end{equation}
\begin{equation}
    \displaystyle
    {\Theta}^{\star} = \arg \min_{\Theta} \sum_{(\mathbf{x}, y_b) \in \mathcal{D}^{train}_{base}}\sum_{j=1}^r \mathcal{L}^{ce}(\mathbf{W}^T f^j_{\theta}(\mathbf{x}) + \mathbf{b}, y_b) + \mathcal{R}(\Theta),
    \label{eq:dc-pretraining}
\end{equation}
where feature map $f_{\theta}(\mathbf{x})$ is regared as a set of $r=h \times w$ local descriptors in $\mathbb{R}^d$ with $f^j_{\theta}(\mathbf{x})$ representing the $j$-th item in set.
% It is natural to assume image $\mathbf{x}$ and its local descriptors share the same label $y_b$.
Though, from the Mathematics point of view, GAP-pretraining in (\ref{eq:gap-pretraining}) and DC-pretraining in (\ref{eq:dc-pretraining}) are almost identical, we experimentally demonstrate that GAP-pretraining would lead to severe semantic discrepancy among local descriptors (see Fig~\ref{fig:distractor}) when maintaining image-level semantics. DC-pretraining can alleviate the problem by forcing to classify each local descriptor correctly, which explicitly enforces semantic consistency among local descriptors.
In practice, we transform the image label $y_b$ into one-hot hard label $\mathbf{y}_b$ and apply label smoothing~\cite{szegedy2016rethinking} with smoothing parameter $\epsilon = 0.1$ to turn it into a soft label $\tilde{\mathbf{y}}_b$ for model training.

\subsection{Meta-finetuning with Attentive Pooling}
\label{sec:metafinetuning}
\begin{figure*}[!tb]
% \vspace{-0.5em}
\begin{center}
    \includegraphics[width=0.99\textwidth]{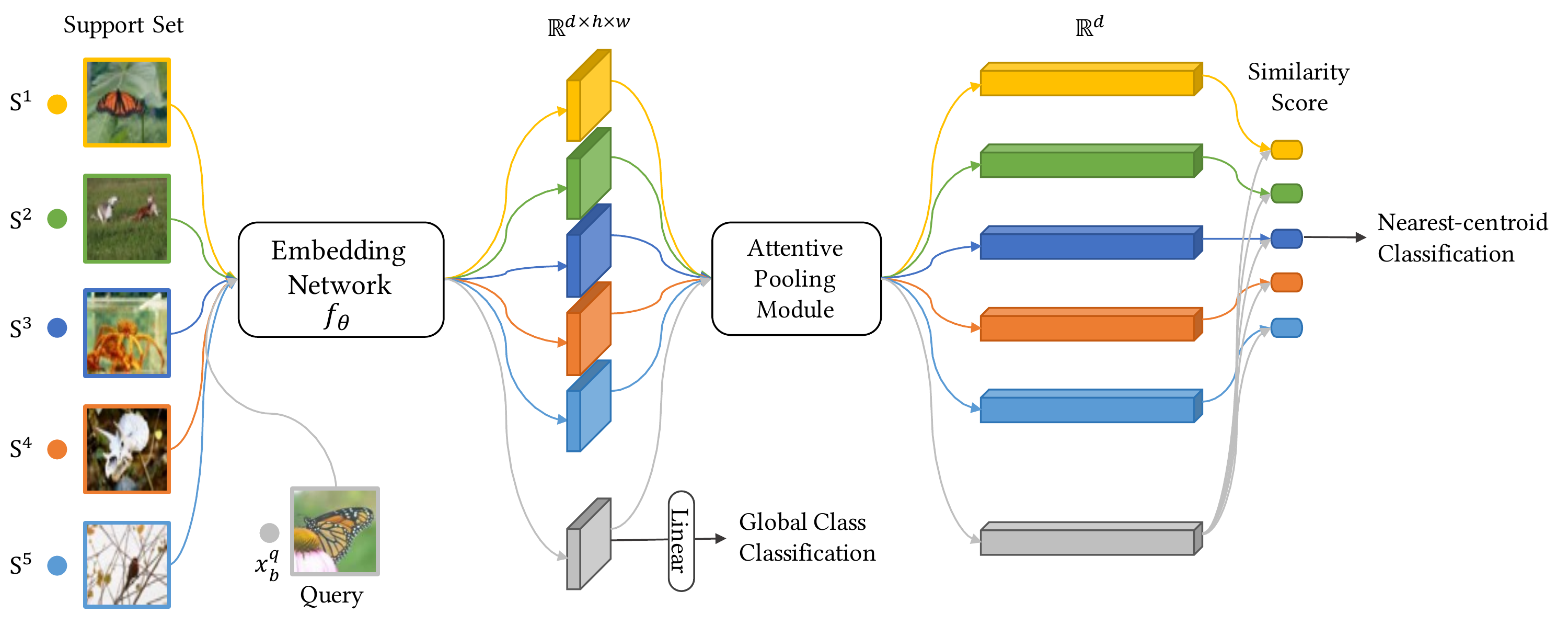}
\end{center}
% \vspace{-1em}
\caption{\textbf{Meta-finetuning for $5$-way $1$-shot setting}. Images are first encoded into feature maps by the pre-trained embedding network, and further shrink into image-level embeddings by applying attentive pooling. Losses from both nearest-centroid classification over image-level embeddings and global class classification over local descriptors are added for model optimization.}
\label{fig:framework}
\end{figure*}
The pre-training in general ends up with a tiny classification loss which indicates the presence of a set of good embeddings.
However, we find that the pre-trained embeddings are not well-prepared for few-shot classification unless an extra meta-finetuning is thereafter performed.
By meta-finetuning, we mean to fine-tune the pre-trained embedding network on the meta-training set $\mathcal{D}_{meta}^{train}$ to perform nearest-centroid few-shot classification following the standard meta-learning paradigm described in \S\ref{sec:fsc_metalearning}.

\begin{figure}[t]
% \vspace{-0.5em}
\begin{center}
    \includegraphics[width=0.68\linewidth]{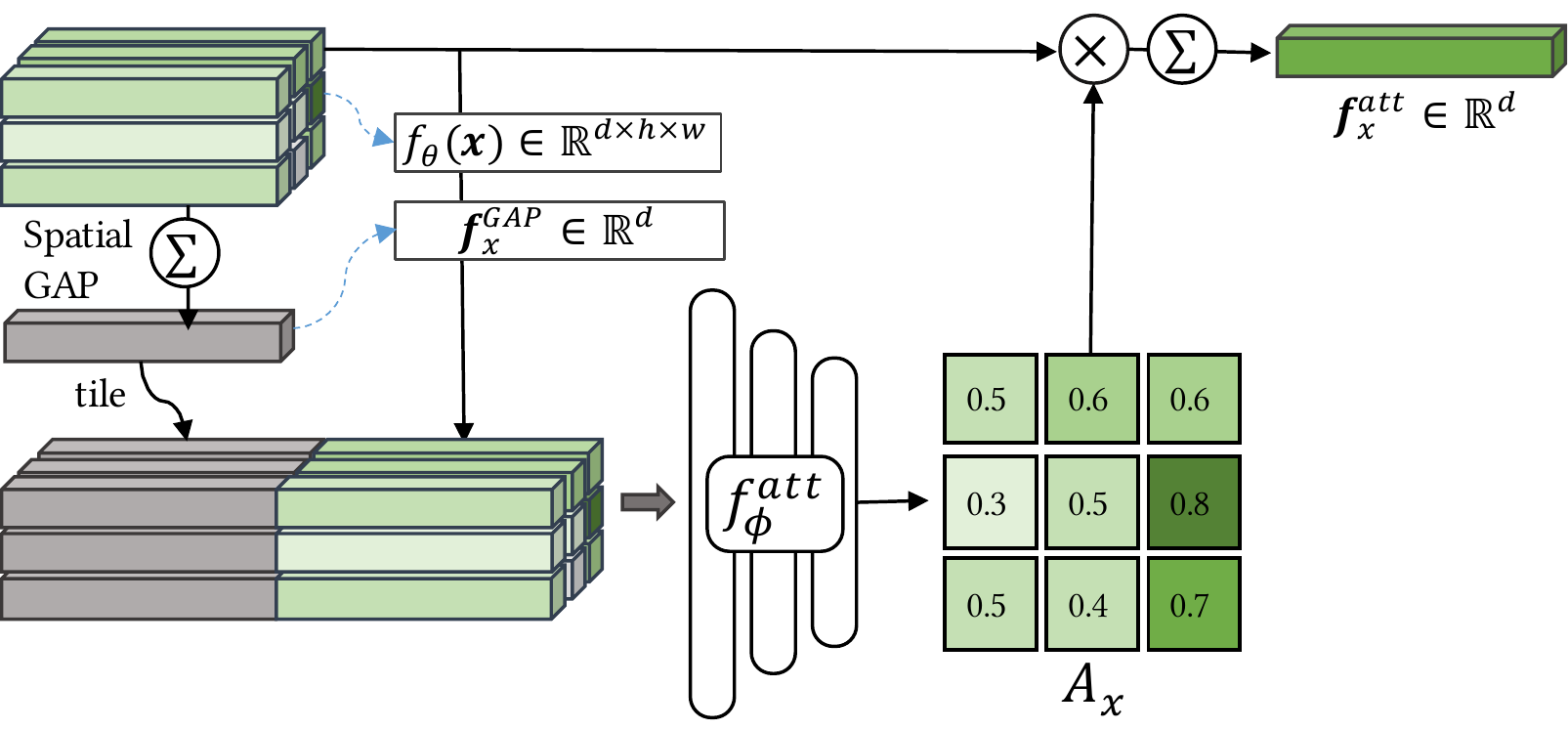}
\end{center}
% \vspace{-1em}
\caption{ \textbf{Attentive pooling module.} Tensor $f_{\theta}(\mathbf{x})$ represents the feature map of input image $\mathbf{x}$, and $\mathbf{f}_x^{GAP}$ denotes its visual representation after applying GAP. The feature map is seen as $h \times w$ local descriptors in $\mathbb{R}^d$ (here $h=3, w=3$) with each representing a local region of the input image. We concatenate each local descriptor and $\mathbf{f}^{GAP}_x$ together before feeding the new generated vectors into regressor $f_{\phi}^{att}$ to estimate region importances. Local descriptors are scaled accordingly and pooled into an image-level embedding $\mathbf{f}_x^{att}$ in $\mathbb{R}^d$ before used for few-shot classification.}
\label{fig:attpool}
\end{figure}

In this meta-training stage, previous work usually apply GAP over feature maps to pool local descriptors in the feature map into compact image-level embeddings for few-shot classification.
We argue that different local descriptors represent different image regions and, as a consequence, they should be of different importances during the pooling operation.
Thus, we propose to reweight local descriptors right before summarizing them into the image-level embedding so that some informative local descriptors in a feature map will surpass the other less informative local descriptors to contribute a more discriminative feature embedding.
As illustrated in Figure~\ref{fig:framework}, given a training task $\mathcal{T}_i = (\mathcal{D}_i^{train}, \mathcal{D}_i^{test})$ where $\mathcal{T}_i \in \mathcal{D}_{meta}^{train}$, images are first encoded into the embedding space, each represented by a $3$-d feature map in $\mathbb{R}^{d \times h \times w}$.
We next summarize each feature map into an image-level embedding in $\mathbb{R}^d$ by applying attentive pooling, which takes the form
% \begin{align}
%     \mathbf{f}_x^{att} 
%     &= AttPool(f_{\theta}(\mathbf{x})) = \sum_{j=1}^{r} {\alpha}_x^j f_{\theta}^j(\mathbf{x}), \nonumber \\
%     & \st~{\alpha}_x^j = \frac{{A}_x^j}{\sum_i^{r}{A}_x^i} 
%     \label{eq:attpool}
% \end{align}
\begin{equation}
    \mathbf{f}_x^{att} = AttPool(f_{\theta}(\mathbf{x})) = \sum_{j=1}^{r} {\alpha}_x^j f_{\theta}^j(\mathbf{x}), ~~~\text{   }~~\st~{\alpha}_x^j = \frac{{A}_x^j}{\sum_i^{r}{A}_x^i} 
    \label{eq:attpool}
\end{equation}
where ${\alpha}_x^j$ denotes the importance coefficient of the $j$-th local descriptor in the feature map.
The attention map ${A}_x$, as illustrated in Figure~\ref{fig:attpool}, is estimated by a simple regressor $f_{\phi}^{att}$ with ${A}_x^i = {\sigma}(f_{\phi}^{att}([\mathbf{f}_x^{GAP}, f_{\theta}^i(\mathbf{x})]))$, where $[\cdot]$ is the concatenation operator and $\sigma$ is the well-known logistic function. 
Finally, we compute the few-shot classification loss on task $\mathcal{T}_i$ as follows
\begin{align}
    &L^{meta} = \frac{-1}{\vert \mathcal{D}_i^{test} \vert} \sum_{(\mathbf{x}, y) \in \mathcal{D}_i^{test}} \log(p_{\mathbf{x}}( \hat{y} = y\vert \mathcal{D}_i^{train})), \nonumber \\
    &~\st~ p_{\mathbf{x}}( \hat{y} = y\vert \mathcal{D}_i^{train}) = \frac{e^{\langle\mathbf{f}_x^{att} , \mathbf{c}_y/{\parallel \mathbf{c}_y \parallel}_2\rangle}}{\sum_{j=1}^N e^{\langle\mathbf{f}_x^{att} , \mathbf{c}_j/{\parallel \mathbf{c}_j \parallel}_2 \rangle}} 
    \label{eq:fs_classifier}
\end{align}
where $p_{\mathbf{x}}( \hat{y} = y\vert \mathcal{D}_i^{train})$ is the probability of query image $\mathbf{x}$ being classified right.
% Subscripts, learnable parameters and the regularization term are omitted in (\ref{eq:fs_classifier}) for brevity. 
Learnable parameters and the regularization term are omitted in (\ref{eq:fs_classifier}) for brevity. 

Intuitively, the attentive pooling explains what the learner is looking for as evidence for decision making for few-shot classification.
By strengthening the activations of some local descriptors while suppressing the activations of the others, it intends to select more informative local descriptors to build better embeddings for classification.
Nevertheless, one challenge we experienced is that the generated attention map ${A}_x$ tends to be quite unevenly distributed, with one or several local descriptors getting much higher attention scores and in turn overwhelming the whole feature map, leading to information loss during pooling.
This is because the attention regressor $f_{\phi}^{att}$ works in an unsupervised manner and is implicitly optimized during the overall learning procedure.
To avoid this issue, we impose an entropy regularization on the normalized attention map, which takes the form
\begin{equation}
    L^{entropy} = \frac{1}{\vert \mathcal{D}_i^{test} \vert}\sum_{\mathbf{x}\in\mathcal{D}_i^{test}}\sum_{j=1}^{r}{\alpha}_x^j\log{{\alpha}_x^j}.
\end{equation}
% This entropy regularization guarantees a smoother attention map. 
The regularization guarantees a smoother attention map. 

It is also worthy to mention that we simultaneously perform global classification over local descriptors alongside few-shot classification in meta-finetuning. It follows that 
\begin{equation}
    L^{ce} = \frac{1}{\vert \mathcal{D}_i^{test} \vert} \sum_{(\mathbf{x}, y_b)\in\mathcal{D}_i^{test}}\sum_{j=1}^{r}\mathcal{L}^{ce}(\mathbf{W}^T f^j_{\theta}(\mathbf{x}) + \mathbf{b}, y_b),
\end{equation}
where parameters $(\mathbf{W}, \mathbf{b})$ are initialized from the pre-trained global classifier and fixed. 
This dense classification encourages local descriptors to maintain their semantics when fine-tuned for few-shot classification, playing a role of knowledge distillation. 
We regularize the global classification with label smoothing too. 
% For input image $(\mathbf{x}, y_b)$, we take its attention map for label smoothing regularization. Specifically, the soft label of a local descriptor $f_{\theta}^j(\mathbf{x})$ is given by setting the smoothing parameter $\epsilon = 1-A_x^j$.  
For input image $(\mathbf{x}, y_b)$, the soft label of a local descriptor $f_{\theta}^j(\mathbf{x})$ is given by setting the smoothing parameter $\epsilon = 1-A_x^j$. 
The combination of the aforementioned meta-training loss, entropy regularization and the global classification loss leads to the overall learning objective of meta-finetuning, which takes the form
\begin{equation}
    \min_{\Omega} E_{\mathcal{T}_i\in\mathcal{D}^{train}_{meta}}(L^{meta} + \beta L^{entropy} + \gamma L^{ce}) + \mathcal{R}(\Omega),
    \label{eq:metafinetune}
\end{equation}
where $\Omega = (\theta, \phi)$ are learnable parameters of both the pre-trained embedding network $f_{\theta}$ and the randomly initialized attention regressor $f_{\phi}^{att}$. $(\beta, \gamma)$ are weights to balance the effects of different losses. As we shall see, the meta-finetuning can quickly adapt a pre-trained model to few-shot setting, much faster and better than meta-training from scratch.

\section{Experiments}
\label{sec:experiment}
In this section, we evaluate the benefits of different components of our proposed DCAP and compare our results with prior arts on two benchmark datasets, \ie \textit{mini}Imagenet~\cite{vinyals2016matching} and \textit{tiered}Imagenet~\cite{ren2018meta}. 

\subsection{Datasets}
\noindent\textbf{\textit{mini}Imagenet.} 
The \textit{mini}Imagenet dataset is a derivative of the larger ILSVRC-12 dataset~\cite{russakovsky2015imagenet}. 
It consists of 60,000 color images of size $84 \times 84$ that are divided into 100 classes with 600 images each. 
We follow \cite{ravi2017optimization,snell2017prototypical} to split the dataset into 64 base classes for meta-training, 16 novel classes for meta-validation, and 20 novel classes for meta-testing, respectively. 

\noindent\textbf{\textit{tiered}Imagenet.} 
The \textit{tiered}Imagenet dataset is another derivative of the ILSVRC-12 dataset. 
It contains more than 700,000 images, divided into 608 classes with the average number of images in each class being more than 1200.
Particularly, it has a hierarchical structure with all its 608 classes derived from 34 high-level nodes in ILSVRC-12. 
The 34 top categories are divided into 20 meta-training (351 classes), 6 meta-validation (97 classes) and 8 meta-testing (160 classes) categories. 
This high-level split provides a more challenging and realistic few-shot setting where the meta-training set is distinctive enough from the meta-testing set semantically. 

% \subsection{Implementation Details}
% \subsection{Experimental Setup}
\subsection{Architecture}
\noindent\textbf{Embedding Network.} We use two backbones as the architecture of our embedding network $f_{\theta}$ for experiments -- Resnet12 and Conv4. 
Resnet12~\cite{he2016deep} contains 4 residual blocks of depth 3 with $3 \times 3$ kernels. 
Convolutional layer depth starts with 64 filters and is doubled every next block. 
A  $2 \times 2$ max pooling is applied at the end of each block for down-sampling. 
% We donot use Dropblock~\cite{ghiasi2018dropblock} for regularization in our implementation.
Likewise, the Conv4 refers to a family of networks that contains 4 convolutional blocks, where each block contains a convolutional layer with $3 \times 3$ kernels, a batch normalization layer, a relu activation layer and a $2 \times 2$ max pooling layer. 
Conv4 networks differs from each other in the number of filters in each convolutional block. For example, the number of filters in Conv4-64 is $(64,64,64,64)$. Tab~\ref{tab:conv4} lists 4 frequently used Conv4 variants. 
% In all experiments, we train the backbones using the SGD optimizer with a momentum of $0.9$ and a weight decay of $0.0005$. 
  
\begin{table}[htb]
\centering
\small
\caption{\textbf{4 frequently used Conv4 variants in the literature.}}
% \vspace{0em}
% \resizebox{1.0\linewidth}{!}{
\setlength{\tabcolsep}{5.5pt}
\begin{tabular}{ lc|cc}
\hline
\toprule
\textbf{name} &&& \textbf{\#filters} \\
\midrule
Conv4-32 &&& $(32,32,32,32)$ \\
Conv4-64 &&& $(64,64,64,64)$ \\
Conv4-128 &&& $(64,64,128,128)$ \\
Conv4-256 &&& $(64,96,128,256)$ \\
\bottomrule
\end{tabular}
% }
% \vspace{-0.5em}
\label{tab:conv4}
\end{table}

\noindent\textbf{Attention Regressor.} The attention regressor $f_{\phi}^{att}$ is a plain convolutional network of depth 2 with 8 hidden units in our implementation. The first convolutional layer reduces the embedding space dimension from $d \times 2$ to $8$, and the second convolutional layer maps embeddings in the latent space $\mathbb{R}^8$ into attention scores. Each attention score is a scalar that represents the importance of the corresponding local descriptor. While more sophisticated structures are applicable, it is beyond the scope of this paper.

\noindent\textbf{Global Classifier.} The global classifier used in dense classification pre-training is a single fully connected layer with no hidden units. This linear layer maps local descriptors in $\mathbb{R}^d$ into logits in $\mathbb{R}^{64}$  (for \textit{mini}Imagenet) or  $\mathbb{R}^{351}$ (for \textit{tiered}Imagenet). A softmax transformation then turns the logits into probability distributions. In meta-finetuning, the auxiliary global classification adopts a global classifier of the same structure.

\subsection{Implementation Details}

\noindent\textbf{Pre-training.}
We train a 64-way global classifier on the meta-training split of \textit{mini}Imagenet and a 351-way global classifier on the meta-training split of \textit{tiered}Imagenet using standard cross-entropy loss.
In this stage, we follow the conventional deep learning pipeline to horizontally divide each meta-training class into two parts, one for model training and the other for evaluation. The sample rate is set to 0.1.
That is, for a meta-training class that has 1000 images in total, we randomly sample 100 images out of the pool for evaluation and use the other 900 images for model training.
The learning rate used in this stage is initialized as $0.1$.
On \textit{mini}Imagenet, we train for 100 epochs with a batch size of 256. The learning rate decays once at 80. 
On \textit{tiered}Imagenet, we train for 120 epochs with a batch size of 512, and accordingly the learning rate decays at 50 and 90.
We adopt different data augmentations when dealing with the two datasets.
Particularly, on \textit{mini}Imagenet, we adopt the random horizontal flip data augmentation only.
On \textit{tiered}Imagenet, both random resized crop and horizontal flip data augmentations are used for model optimization.

\noindent\textbf{Meta-fintuning.} 
The pre-trained embedding network is fine-tuned at a learning rate of $0.001$ on the meta-training set that contains $20k$/$50k$ tasks randomly sampled from the meta-training split of \textit{mini}Imagenet/\textit{tiered}Imagenet for few-shot classification.
During meta-finetuning, the attention regressor is randomly initialized and optimized with an initial learning rate of $0.01$.
We follow prior art~\cite{hou2019cross,lee2019meta,ye2020few} to consider both \textit{5-way 1-shot} and \textit{5-way 5-shot} few-shot settings.
In \textit{5-way 1-shot} setting, a few-shot task has in total 80 images, where each class contains one support image and 15 query images. 
In \textit{5-way 5-shot} setting, the number of query images per class decreases to 10 so as to fit in well with the GPU memory.
We consider a mini-batch that consists of 4 few-shot tasks and adopt the same data augmentation (\ie random horizontal flip) when dealing with both datasets. 
The learning rate decays once at $16k$ and $40k$ on \textit{mini}Imagenet and \textit{tiered}Imagenet respectively.

\noindent\textbf{Evaluation Protocols.} 
To be consistent with previous work~\cite{he2020memory,tian2020rethinking,simon2020adaptive,zhang2020uncertainty}, we construct a meta-evaluation/meta-testing set by randomly sampling $1000$ few-shot tasks from the meta-validation/meta-testing split of each dataset for model validation/testing. 
Mean accuracy along with the $95\%$ confidence interval over the $1000$ tasks is reported for model comparison. 
Besides, we follow~\cite{chen2020new} to apply \textit{consistent sampling} when preparing the meta-evaluation/meta-testing set. Consistent sampling ensures that the same set of tasks are sampled for model evaluation/testing in each epoch following a deterministic order. 

All our experiments are implemented in PyTorch\footnote{PyTorch: \url{https://pytorch.org}} on an Ubuntu server with 4 NVIDIA TITAN Xp GPU cards.
In both pre-training and meta-finetuning, we adopt SGD with Nesterov momentum of $0.9$ and weight decay of $0.0005$ for model optimization. 
The learning rate decay is set to $0.1$ in two stages. 
Unless otherwise specified, we set $\beta=0.1$ and $\gamma=0.5$ in (\ref{eq:metafinetune}) in all experiments.

\subsection{Results with ResNet12 Backbone}
\label{sec:result_resnet12}

\begin{table*}[!t]
% \vspace{0em}
\caption{
\textbf{Comparison to prior work on \textit{mini}ImageNet and \textit{tiered}ImageNet (ResNet12).} Mean accuracies (\%) with 95\% confidence intervals on \textit{mini}ImageNet and \textit{tiered}ImageNet meta-testing set. 
% ResNet-12\textsuperscript{$\ast$} denotes a $1.25\times$ wider version of ResNet-12. 
% ResNet-12\textsuperscript{$\diamond$} denotes \textit{ResNet-256F}, a narrower version of ResNet12. 
\textsuperscript{$\dagger$} models use transductive inference. 
% $*$-trainval results are obtained by training on the union of meta-training and meta-validation splits.
\textsuperscript{$\ast$} Results from~\cite{simon2020adaptive}. 
\textsuperscript{$\ddagger$} refers to our reimplementation.
}
\begin{center}
% \resizebox{0.96\linewidth}{!}{

\begin{small}
\begin{tabular}{@{}llc@{}cc@{}c@{}cc@{}}
\hline
\toprule
& & \phantom{a} & \multicolumn{2}{c}{\textbf{\textit{mini}ImageNet 5-way}} & \phantom{ab} & \multicolumn{2}{c}{\textbf{\textit{tiered}ImageNet 5-way}} \\
\cmidrule{4-5} \cmidrule{7-8}
\textbf{model} & \textbf{backbone} && \textbf{1-shot} & \textbf{5-shot} && \textbf{1-shot} & \textbf{5-shot}  \\
\midrule

Prototypical Networks\textsuperscript{$\ast$} \cite{snell2017prototypical} & ResNet-12 && 59.25 $\pm$ 0.64 & 75.60 $\pm$ 0.48 && 61.74 $\pm$ 0.77 & 80.00 $\pm$ 0.55 \\
Prototypical Networks\textsuperscript{$\ddagger$}~\cite{snell2017prototypical} & ResNet-12 && 59.25 $\pm$ 0.70 & 76.88 $\pm$ 0.59 && 62.15 $\pm$ 0.71 & 81.72 $\pm$ 0.53 \\
% SNAIL~\cite{mishra2018snail} & ResNet-12\textsuperscript{$\diamond$} && 55.71 $\pm$ 0.99 & 68.88 $\pm$ 0.92 && - & - \\
SNAIL~\cite{mishra2018snail} & ResNet-12 && 55.71 $\pm$ 0.99 & 68.88 $\pm$ 0.92 && - & - \\
TADAM~\cite{oreshkin2018tadam} & ResNet-12 && 58.50 $\pm$ 0.30 & 76.70 $\pm$ 0.30 && - & - \\
% DN4~\cite{li2019revisiting} & ResNet-12\textsuperscript{$\diamond$} && 54.37 $\pm$ 0.36 & 74.44 $\pm$ 0.29 && - & - \\
DN4~\cite{li2019revisiting} & ResNet-12 && 54.37 $\pm$ 0.36 & 74.44 $\pm$ 0.29 && - & - \\
TPN\textsuperscript{$\dagger$}~\cite{liu2019learning}  & ResNet-12 && 59.46 $\pm$ - & 75.65 $\pm$ - && - & - \\
Fine-tuning~\cite{dhillon2019baseline} & ResNet-12 && 56.67 $\pm$ 0.62 & 74.70 $\pm$ 0.51 && 64.45 $\pm$ 0.70 & 83.59 $\pm$ 0.51 \\
Fine-tuning\textsuperscript{$\dagger$}~\cite{dhillon2019baseline} & ResNet-12 && 62.35 $\pm$ 0.66 & 74.53 $\pm$ 0.54 && 68.41 $\pm$ 0.73 & 83.41 $\pm$ 0.52 \\
MTL~\cite{sun2019meta} & ResNet-12 && 61.20 $\pm$ 1.80 & 75.50 $\pm$ 0.80 && -  & - \\
DC~\cite{lifchitz2019dense} & ResNet-12 && 62.53 $\pm$ 0.19 & 78.95 $\pm$ 0.13 && -  & - \\
CAN~\cite{hou2019cross} & ResNet-12 && 63.85 $\pm$ 0.48 & 79.44 $\pm$ 0.34 && \textbf{69.89} $\pm$ 0.51 & 84.23 $\pm$ 0.37 \\
FEAT\textsuperscript{$\ast$}~\cite{ye2020few} & ResNet-12 && 61.72 $\pm$ 0.11 & 78.32 $\pm$ 0.76 && -  & - \\
Meta-Baseline~\cite{chen2020new} & ResNet-12 && 63.17 $\pm$ 0.23 & 79.26 $\pm$ 0.17 && 68.62 $\pm$ 0.27 & 83.29 $\pm$ 0.18 \\
% MetaOptNet~\cite{lee2019meta} & ResNet-12\textsuperscript{$\ast$} && 62.64 $\pm$ 0.61 & 78.63 $\pm$ 0.46 && 65.99 $\pm$ 0.72 & 81.56 $\pm$ 0.53 \\
MetaOptNet~\cite{lee2019meta} & ResNet-12 && 62.64 $\pm$ 0.61 & 78.63 $\pm$ 0.46 && 65.99 $\pm$ 0.72 & 81.56 $\pm$ 0.53 \\
% RFS~\cite{tian2020rethinking} & ResNet-12\textsuperscript{$\ast$} && 62.02 $\pm$ 0.63 & 79.64 $\pm$ 0.44 && 69.74 $\pm$ 0.72 & 84.41 $\pm$ 0.55 \\
RFS~\cite{tian2020rethinking} & ResNet-12 && 62.02 $\pm$ 0.63 & 79.64 $\pm$ 0.44 && 69.74 $\pm$ 0.72 & \textbf{84.41} $\pm$ 0.55 \\
% RFS-distill\cite{tian2020rethinking} & ResNet-12\textsuperscript{$\ast$} && 64.82 $\pm$ 0.60 & 82.14 $\pm$ 0.43 && 71.52 $\pm$ 0.69 & 86.03 $\pm$ 0.49 \\
% DSN~\cite{simon2020adaptive} & ResNet-12\textsuperscript{$\ast$} && 62.64 $\pm$ 0.66 & 78.83 $\pm$ 0.45 && 66.22 $\pm$ 0.75 & 82.79 $\pm$ 0.48 \\
DSN~\cite{simon2020adaptive} & ResNet-12 && 62.64 $\pm$ 0.66 & 78.83 $\pm$ 0.45 && 66.22 $\pm$ 0.75 & 82.79 $\pm$ 0.48 \\
Meta-UAFS~\cite{zhang2020uncertainty} & ResNet-12 && \textbf{64.22} $\pm$ 0.67 & \textbf{79.99} $\pm$ 0.49 && 69.13 $\pm$ 0.84 & 84.33 $\pm$ 0.59 \\

\midrule
\textbf{DCAP (ours)} & ResNet-12 && \textbf{65.20} $\pm$ 0.67 & \textbf{80.93} $\pm$ 0.53 && \textbf{70.15} $\pm$ 0.74 & \textbf{85.33} $\pm$ 0.55\\
\bottomrule
\hline
\end{tabular}
\end{small}
% }
\end{center}

% \vspace{-0.5em}
% \vspace{0em}
\label{tab:result_resnet12}
\end{table*}
  
Tab~\ref{tab:result_resnet12} compares our method with other few-shot algorithms when ResNet12 is used as the backbone network. 
As it can be seen, our proposed DCAP can consistently perform better than prior art on \textit{mini}Imagenet and \textit{tiered}Imagenet under both \textit{5-way 1-shot} and \textit{5-way 5-shot} settings.
Specifically, our method is around $3\%/6.4\%$, $4\%/5.4\%$, $2.7\%/2\%$, $1.4\%/1.5\%$, $3.5\%/2.6\%$, $2\%/1.7\%$, $2.6\%/2.3\%$, $3.2\%/1.3\%$, $2.6\%/2.1\%$, $1\%/0.9\%$ better than Fine-tuning~\cite{dhillon2019baseline}, MTL~\cite{sun2019meta}, DC~\cite{lifchitz2019dense}, CAN~\cite{hou2019cross}, FEAT~\cite{ye2020few}, Meta-Baseline~\cite{chen2020new}, MetaOptNet~\cite{lee2019meta}, RFS~\cite{tian2020rethinking}, DSN~\cite{simon2020adaptive}, Meta-UAFS~\cite{zhang2020uncertainty} under the \textit{1-shot}/\textit{5-shot} setting on \textit{mini}Imagenet, respectively.
On \textit{tiered}Imagenet, our method is around $1.7\%/2\%$, $0.2\%/1\%$, $1.5\%/2\%$, $4.1\%/3.8\%$, $0.4\%/0.9\%$, $3.9\%/2.6\%$, $1\%/1\%$ better than Fine-tuning~\cite{dhillon2019baseline}, CAN~\cite{hou2019cross}, Meta-Baseline~\cite{chen2020new}, MetaOptNet~\cite{lee2019meta}, RFS~\cite{tian2020rethinking}, DSN~\cite{simon2020adaptive}, Meta-UAFS~\cite{zhang2020uncertainty}, respectively.
Notably, we observe that Prototypical Networks provide a strong baseline. It not only outperforms SNAIL~\cite{mishra2018snail} and DN4~\cite{li2019revisiting} by a large margin, but also has comparable results against TADAM~\cite{oreshkin2018tadam} and even TPN~\cite{liu2019learning}, a transductive method in which all query images arrive simultaneously during inference.
In principle, transductive methods are able to achieve higher performances than inductive methods because the transductive methods can leverage information in other query images when dealing with the current query image while the inductive methods can not. 

Note that, RFS and DSN reported in Tab~\ref{tab:result_resnet12} are not the best models as reported in the original papers.
In~\cite{tian2020rethinking}, RFS exploited Born-again~\cite{furlanllo2018born} strategy to apply knowledge distillation sequentially to revise the embedding network and its associated base learner multiple times. The strategy is not adopted by other works, including ours of course. For fair comparison, we report the results of its variant without Born-again strategy in the table.
DSN~\cite{simon2020adaptive} proposed to refine the prototype of each class by taking advantage of unlabeled data when addressing semi-supervised few-shot classification problem and achieved outstanding performance. In this work, we consider the standard few-shot classification setting but leave the semi-supervised setting untouched. Therefore, we also choose to report results of its variant that studies the same standard few-shot classification for fair comparison.

\subsection{Results with Conv4 Backbone}
\label{sec:result_conv4}

\begin{table*}[!t]
% \vspace{0em}
\caption{
\textbf{Comparison to prior work on \textit{mini}ImageNet and \textit{tiered}ImageNet (Conv4).} 
Mean accuracies (\%) with 95\% confidence intervals on \textit{mini}ImageNet and \textit{tiered}ImageNet meta-testing set. 
a-b-c-d denotes a Conv4 network with a, b, c, and d filters in each convolutional block. 
\textsuperscript{$\dagger$} models use transductive inference.
\textsuperscript{$\ddagger$} refers to our reimplementation.}
\begin{center}
% \resizebox{0.96\linewidth}{!}{

\begin{small}
\begin{tabular}{@{}llc@{}cc@{}c@{}cc@{}}
\hline
\toprule
& & \phantom{a} & \multicolumn{2}{c}{\textbf{\textit{mini}ImageNet 5-way}} & \phantom{ab} & \multicolumn{2}{c}{\textbf{\textit{tiered}ImageNet 5-way}} \\
\cmidrule{4-5} \cmidrule{7-8}
\textbf{model} & \textbf{backbone} && \textbf{1-shot} & \textbf{5-shot} && \textbf{1-shot} & \textbf{5-shot}  \\
\midrule
MAML~\cite{finn2017model} & 32-32-32-32 &&  48.70 $\pm$ 1.84 & 63.11 $\pm$ 0.92 && 51.67 $\pm$ 1.81 & 70.30 $\pm$ 1.75 \\
Meta-SGD~\cite{li2017metasgd} & 32-32-32-32 && 50.47 $\pm$ 1.87 & 64.03 $\pm$ 0.94 && - & - \\
Meta-Learner~\cite{ravi2017optimization} & 32-32-32-32 && 43.44 $\pm$ 0.77 & 60.60 $\pm$ 0.71 && - & - \\
Reptile~\cite{nichol2018reptile} & 32-32-32-32 && 49.97 $\pm$ 0.32 & 65.99 $\pm$ 0.58 && - & - \\
TAML~\cite{jamal2019task} & 32-32-32-32 && 49.33 $\pm$ 1.8 & 66.05 $\pm$ 0.85 && - & - \\

Matching Networks~\cite{vinyals2016matching} & 64-64-64-64 && 43.56 $\pm$ 0.84 & 55.31 $\pm$ 0.73 && - & - \\
Prototypical Networks~\cite{snell2017prototypical} & 64-64-64-64 && 46.14 $\pm$ 0.77 & 65.77 $\pm$ 0.70 && 48.58 $\pm$ 0.87 & 69.57 $\pm$ 0.75 \\
Prototypical Networks\textsuperscript{$\ddagger$}~\cite{snell2017prototypical} & 64-64-64-64 && 49.76 $\pm$ 0.60 & 67.63 $\pm$ 0.58 && 51.32 $\pm$ 0.68 & 70.33 $\pm$ 0.59 \\
Relation Networks \cite{sung2018learning} & 64-64-64-64 && 51.38 $\pm$ 0.82 & 67.07 $\pm$ 0.69 && 54.48 $\pm$ 0.93 & 71.31 $\pm$ 0.78 \\
DN4~\cite{li2019revisiting} & 64-64-64-64 && 51.24 $\pm$ 0.74 & 71.02 $\pm$ 0.64 && - & - \\
MetaOptNet~\cite{lee2019meta} & 64-64-64-64 && 52.87 $\pm$ 0.57 & 68.76 $\pm$ 0.48 && 54.71 $\pm$ 0.67 & 71.79 $\pm$ 0.59 \\
PARN~\cite{wu2019parn} & 64-64-64-64 && \textbf{55.22} $\pm$ 0.84 & \textbf{71.55} $\pm$ 0.66 && - & - \\
Fine-tuning~\cite{dhillon2019baseline} & 64-64-64-64 && 49.43 $\pm$ 0.62 & 66.42 $\pm$ 0.53 && \textbf{57.45} $\pm$ 0.68 & \textbf{73.96} $\pm$ 0.56 \\
Fine-tuning\textsuperscript{$\dagger$}~\cite{dhillon2019baseline} & 64-64-64-64 && 50.46 $\pm$ 0.62 & 66.68 $\pm$ 0.52 && 58.05 $\pm$ 0.68 & \textbf{74.24} $\pm$ 0.56 \\
TPN\textsuperscript{$\dagger$}~\cite{liu2019learning}  & 64-64-64-64 && 53.75 $\pm$ 0.86 & 69.43 $\pm$ 0.67 && 57.53 $\pm$ 0.96 & 72.85 $\pm$ 0.74 \\
MRN\textsuperscript{$\dagger$}~\cite{he2020memory} & 64-64-64-64 && \textbf{57.83} $\pm$ 0.69 & \textbf{71.13} $\pm$ 0.50 && \textbf{62.65} $\pm$ 0.84 & 74.20 $\pm$ 0.64 \\
DSN~\cite{simon2020adaptive} & 64-64-64-64 && 51.78 $\pm$ 0.96 & 68.99 $\pm$ 0.69 && - & - \\

GNN~\cite{satorras2018gnn} & 64-96-128-256 && 50.33 $\pm$ 0.36 & 66.41 $\pm$ 0.63 && - & - \\

\midrule
\textbf{DCAP (ours)} & 64-64-64-64 && 53.68 $\pm$ 0.63 & 68.62 $\pm$ 0.60 && 55.63 $\pm$ 0.73 & 70.78 $\pm$ 0.60\\
& 64-64-128-128 && 55.25 $\pm$ 0.61 & 70.46 $\pm$ 0.60 && 58.75 $\pm$ 0.73 & 72.81 $\pm$ 0.61\\
\bottomrule
\hline
\end{tabular}
\end{small}
% }
\end{center}

% \vspace{-0.5em}
% \vspace{0em}
\label{tab:results_conv4}
\end{table*}

To give an overview of the literature, we additionally take a Conv4 as the backbone network, evaluate our method on \textit{mini}Imagenet and \textit{tiered}Imagenet under both \textit{5-way 1-shot} and \textit{5-way 5-shot} settings, and compare our method with other works that adopt the same or similar backbone in Tab~\ref{tab:results_conv4}.
Particularly, we follow~\cite{snell2017prototypical,sung2018learning,he2020memory,simon2020adaptive} to adopt the widely used Conv4-64 as described in Tab~\ref{tab:conv4} for experiments. The Conv4-64 maps an input image of size $84 \times 84$ into a feature map of size $64 \times 5 \times 5$.
As can be seen, our method can achieve competitive or better performance when compared to most of the counterparts. 
For example, our method is around $10\%/13\%$, $2.3\%/1.5\%$, $4\%/2.2\%$, $3.3\%/2.2\%$ better than MAML~\cite{finn2017model}, Relation Networks \cite{sung2018learning}, Fine-tuning~\cite{dhillon2019baseline}, GNN~\cite{satorras2018gnn} under \textit{1-shot}/\textit{5-shot} setting on \textit{mini}Imagenet, respectively.
We notice that our method is a little bit inferior to prior art DN4~\cite{li2019revisiting}, MetaOptNet~\cite{lee2019meta} and DSN~\cite{simon2020adaptive} in \textit{5-shot} setting.
It is also observed that our method underperforms transductive methods, like TPN~\cite{liu2019learning} and Fine-tuning~\cite{dhillon2019baseline}, especially on \textit{tiered}Imagenet, which is inconsistent with our observation in \S\ref{sec:result_resnet12}.
We argue the reason of the inferiority is two-fold.
\begin{itemize}
\item First, dense classification pre-training the light-weight Conv4-64 backbone is more difficult because of its limited representation ability. In the pre-training stage, by training the Conv4-64 backbone on the meta-training split of \textit{mini}Imagenet/\textit{tiered}Imagenet for dense classification, we can merely get a top-1 evaluation accuracy of $33.08\% / 12.21\%$. In contrast, pre-training the ResNet12 backbone in this stage gives a top-1 evaluation accuracy of $64.57\% / 56.88\%$. 
The poorly pre-trained backbone explains why the superiority of our method is not preserved when switching from ResNet12 to Conv4-64.
Even so, we are surprised to observe that the pre-trained backbone can achieve results that are not bad, or even better than previous work in \textit{1-shot} setting.
\item Second, a 64-dimensional feature embedding is much less representative than a 1600-dimensional one. A common practice adopted by prior arts when working with Conv4-64 is to flatten feature maps in $\mathcal{R}^{64 \times 5 \times 5}$ to construct 1600-dimensional feature embeddings and use the high-dimensional embeddings in the final classifier for few-shot classification. 
The resulting embeddings keep all information that is originally available in the feature maps, including spatial information and semantic information, leading to no information loss.
In this work, we provide an alternative way to build embeddings by applying attentive pooling over the feature maps, which attentively pools feature maps into 64-dimensional embeddings. 
While attentive pooling can select to emphasize important local descriptors, the pooling, like GAP, would inevitably result in information loss.
As a result, using the 1600-dimensional embeddings for few-shot classification unquestionably gains superior results in some cases. 
\end{itemize} 
To validate our assumptions, we use a larger Conv4-128 backbone for experiments and find that an average of $2\%$ improvement can be obtained on all settings.

\subsection{Cross-Domain Few-Shot Classification}
We further justify the effectiveness of our proposed method under a more challenging cross-domain few-shot classification setting, where the learner is trained and tested on few-shot tasks collected from different domains. To be specific, we follow \cite{chen2019closer,baik2020alfa,wertheimer2021reconstruction} to consider the \textit{miniImagenet $\rightarrow$ CUB} cross-domain scenario, where the learner is trained on tasks sampled from the \textit{mini}Imagenet dataset but evaluated on tasks sampled from the CUB-200-2011 dataset~\cite{wah2011cub}. The CUB-200-2011 dataset (also referred as CUB) contains 11,788 images from 200 bird classes and is originally constructed to serve fine-grained bird classification. In few-shot setting, the CUB dataset is divided into 100 classes for meta-training, 50 classes for meta-validation and 50 classes for meta-testing. We randomly sample 1,000 few-shot tasks from its meta-testing split for cross-domain evaluation. Results are reported in Tab~\ref{tab:results_xdomain}. As can be seen, few-shot classification approaches normally suffer significant performance drop in the presence of domain shift between source and target datasets. Nonetheless, our method keeps outperforming many other participants in both 5-way 1-shot and 5-way 5-shot cross-domain few-shot settings.

\begin{table*}[!t]
% \vspace{0em}
\caption{\textbf{Performance comparison in the cross-domain setting: \textit{mini}Imagenet $\rightarrow$ CUB}. Experiments are performed with both two types of backbones. Symbols and organization match Tab~\ref{tab:results_conv4}. "+FT" indicates the model is fine-tuned towards the target task in evaluation.}  
\begin{center}
% \resizebox{0.96\linewidth}{!}{

\begin{small}
\begin{tabular}{@{}llc@{}cc@{}}
\hline
\toprule
& & \phantom{a} & \multicolumn{2}{c}{\textbf{\textit{mini}ImageNet $\rightarrow$ CUB}}\\
\cmidrule{4-5}
\textbf{model} & \textbf{backbone} && \textbf{5-way 1-shot} & \textbf{5-way 5-shot} \\
\midrule
MAML~\cite{finn2017model,chen2019closer} & ResNet-18 && - & 51.34 $\pm$ 0.72 \\
Prototypical Networks~\cite{snell2017prototypical,chen2019closer} & ResNet-18 && - & 62.02 $\pm$ 0.70\\
Baseline~\cite{chen2019closer} & ResNet-18 && - & 65.57 $\pm$ 0.70 \\
Baseline++~\cite{chen2019closer} & ResNet-18 && - & 62.04 $\pm$ 0.76 \\
Matching Networks+FT~\cite{tseng2020cross,wertheimer2021reconstruction} & ResNet-10 && 36.61 $\pm$ 0.53 & 55.23 $\pm$ 0.83 \\
Relation Networks+FT~\cite{tseng2020cross,wertheimer2021reconstruction} & ResNet-10 && 44.07 $\pm$ 0.77 & 59.46 $\pm$ 0.71 \\
GNN+FT~\cite{tseng2020cross,wertheimer2021reconstruction} & ResNet-10 && \textbf{47.47} $\pm$ 0.75 & \textbf{66.98} $\pm$ 0.68 \\
MetaOptNet~\cite{lee2019meta,mangla2020chart,wertheimer2021reconstruction} & ResNet-12 && 44.79 $\pm$ 0.75 & 64.98 $\pm$ 0.68 \\
MAML+L2F~\cite{baik2020alfa} & ResNet-12 && - & 62.12 $\pm$ 0.21 \\
MAML+ALFA~\cite{baik2020alfa} & ResNet-12 && - & 61.22 $\pm$ 0.22 \\
ALFA+Random Init~\cite{baik2020alfa} & ResNet-12 && - & 60.13 $\pm$ 0.23 \\
ALFA+MAML+L2F~\cite{baik2020alfa} & ResNet-12 && - & 63.24 $\pm$ 0.22 \\
\midrule
\textbf{DCAP (ours)} & ResNet-12 && \textbf{49.65} $\pm$ 0.70 & \textbf{68.25} $\pm$ 0.65 \\
\midrule
MAML~\cite{finn2017model,snell2021bayesian} & 32-32-32-32 &&  34.01 $\pm$ 1.25 & 48.83 $\pm$ 0.62 \\
Bayesian MAML~\cite{yoon2018bayesian,snell2021bayesian} & 32-32-32-32 && 33.52 $\pm$ 0.36 & 51.35 $\pm$ 0.16 \\
Matching Networks~\cite{vinyals2016matching,snell2021bayesian} & 64-64-64-64 && 36.98 $\pm$ 0.06 & 50.72 $\pm$ 0.36\\
Prototypical Networks~\cite{snell2017prototypical,snell2021bayesian} & 64-64-64-64 && 33.27 $\pm$ 1.09 & \textbf{52.16} $\pm$ 0.17\\
Relation Networks~\cite{sung2018learning,snell2021bayesian} & 64-64-64-64 && \textbf{37.13} $\pm$ 0.20 & 51.76 $\pm$ 1.48\\
\midrule
\textbf{DCAP (ours)} & 64-64-64-64 && \textbf{39.66} $\pm$ 0.54 & \textbf{52.22} $\pm$ 0.53 \\
\bottomrule
\hline
\end{tabular}
\end{small}
% }
\end{center}

% \vspace{-0.5em}
% \vspace{0em}
\label{tab:results_xdomain}
\end{table*}

\subsection{Ablation Study}
\begin{table}[tb]
    % \caption{\textbf{\textit{5-way 1-shot} accuracy on \textit{mini}Imagenet.} In pre-training, ``GAP'' refers to standard classification upon image-level embeddings generated by applying GAP. ``DC'' refers to dense classification over local descriptors. In meta-learning, ``GAP'' and ``AttPool'' stand for global average pooling and attentive pooling, respectively. We report the results of meta-training from scratch in the first row.}
    \caption{\textbf{5-way 1-shot accuracy on \textit{mini}Imagenet.} In pre-training, ``GAP'' refers to standard classification upon image-level embeddings generated by applying GAP. ``DC'' refers to dense classification over local descriptors. In meta-learning, ``GAP'' and ``AttPool'' stand for global average pooling and attentive pooling, respectively. We report the results of meta-training from scratch in the first row.}
    \centering
    \small
    % \resizebox{1.0\linewidth}{!}{ % resizebox begin
    \setlength{\tabcolsep}{5.5pt}
    \begin{tabular}{ cccc|ccc}
    \hline
    \toprule
    \multicolumn{3}{c}{\textbf{Stage 1: Pre-training}} && \multicolumn{3}{c}{\textbf{Stage 2: Meta-learning}} \\
    \textbf{w/pre-training} & \textbf{GAP} & \textbf{DC} &&& \textbf{GAP} & \textbf{AttPool} \\
    \midrule
    & &  &&& 59.25 $\pm$ 0.70 & 56.69 $\pm$ 0.68 \\
    \checkmark & \checkmark & &&& 63.58 $\pm$ 0.66 & 63.57 $\pm$ 0.66 \\
    \checkmark & & \checkmark &&& 64.13 $\pm$ 0.69 & 65.20 $\pm$ 0.67 \\
    \bottomrule
    \end{tabular}
    % } % resizebox end
    \label{tab:ablation_twostagelearning}
\end{table}

\begin{table}[t]
    % \caption{\textbf{\textit{5-way 1-shot} accuracy on \textit{mini}Imagenet without meta-finetuning.} Results of models with attentive pooling are not given because the attention regressor is not trained yet.}
    \caption{\textbf{5-way 1-shot accuracy on \textit{mini}Imagenet without meta-finetuning.} Results of models with attentive pooling are not given because the attention regressor is not trained yet.}
    \centering
    \small
    % \vspace{0em}
    % \resizebox{1.0\linewidth}{!}{ % resizebox begin
    \setlength{\tabcolsep}{5.5pt}
    \begin{tabular}{ cccc|ccc}
    \hline
    \toprule
    \multicolumn{3}{c}{\textbf{Stage 1: Pre-training}} && \multicolumn{3}{c}{\textbf{}} \\
    \textbf{w/pre-training} & \textbf{GAP} & \textbf{DC} &&& \textbf{GAP} & \textbf{AttPool} \\
    \midrule
    & &  &&& 26.50 $\pm$ 0.46 & 26.21 $\pm$ 0.46 \\
    % \checkmark & \checkmark &  &&& 20.14 $\pm$ 0.30 & 20.24 $\pm$ 0.32 \\
    \checkmark & \checkmark &  &&& 20.14 $\pm$ 0.30 & - \\
    % \checkmark & & \checkmark &&& 20.14 $\pm$ 0.08 & 19.89 $\pm$ 0.28 \\
    \checkmark & & \checkmark &&& 20.14 $\pm$ 0.08 & - \\
    \bottomrule
    \end{tabular}
    % } % resizebox end
    \label{tab:ablation_metafinetuning}
\end{table}

To better understand our method, we analyze DCAP and the following ablated variants on the \textit{mini}Imagenet dataset with ResNet12 backbone.
\begin{itemize}
    % \item \textbf{DC-AttPool.} 
    \item \textbf{DC-GAP} refers to the model that is trained to solve dense classification in pre-training. In meta-finetuning, it pools feature maps into embeddings by applying GAP.
    \item \textbf{GAP-AttPool} refers to the model that is trained to solve standard classification in pre-training where feature maps are pooled into embeddings by applying GAP. In meta-finetuning, it pools feature maps into embeddings by applying attentive pooling.
    \item \textbf{GAP-GAP}, similar to DC-GAP and GAP-AttPool, is trained to solve standard classification in pre-training and applies GAP over feature maps in meta-finetuning. It shares the same pipeline with \cite{chen2020new}.
    \item \textbf{Zero-GAP} refers to the model that is meta-trained from scratch where feature maps are pooled into embeddings by applying GAP. 
    \item \textbf{Zero-AttPool} similarly denotes the model that is meta-trained from scratch where feature maps are pooled into embeddings by applying attentive pooling.
\end{itemize}
We follow the naming scheme to give our DCAP the alias \textbf{DC-AttPool}. 
The results of different models under \textit{1-shot} setting are reported in Tab~\ref{tab:ablation_twostagelearning} for straightforward comparisons.

\noindent\textbf{Is pre-training stage necessary?} 
% We organize the results of different models in \textit{1-shot} setting in Tab~\ref{tab:ablation_twostagelearning} to present straightforward comparisons.
% Tab~\ref{tab:ablation_twostagelearning} presents the results of different models in \textit{1-shot} setting.
% The results of models that are meta-trained from scratch without pre-training are shown in the first row.
% As can be seen, meta-learning on top of a pre-trained model can bring $4.3\%\sim8.5\%$ improvements over models meta-trained from scratch, which demonstrates the importance of pre-training in few-shot classification. 
As shown in Tab~\ref{tab:ablation_twostagelearning}, meta-learning on top of a pre-trained model can bring $4.3\%\sim8.5\%$ improvements over models meta-trained from scratch, which demonstrates the importance of pre-training in few-shot classification. 
Concretely, the GAP-GAP/DC-GAP model is about $4.3\%/4.9\%$ better than the Zero-GAP model, and the DC-GAP/DC-AttPool model is around $6.9\%/8.5\%$ better than the Zero-AttPool model.
We argue the pre-training provides a model with strong representation ability, whereas the beneficial ability is not fully accessible by straight meta-training on few-shot tasks.
Since the meta-training focuses more on transferability, \ie the generalization ability across tasks, a model that starts from a randomly initialized state would struggle to balance the representation ability and the transferability during meta-training, leading to inferior performance.
From Tab~\ref{tab:ablation_twostagelearning}, our another observation is that models with attentive pooling are more sensitive to pre-training.
We think this is due to the difficulty of optimizing the attention regressor for the selection of informative local descriptors if the model is meta-trained from scratch.

\begin{figure}[t]
\resizebox{0.8\linewidth}{!}{ % resizebox begin
    \includegraphics[]{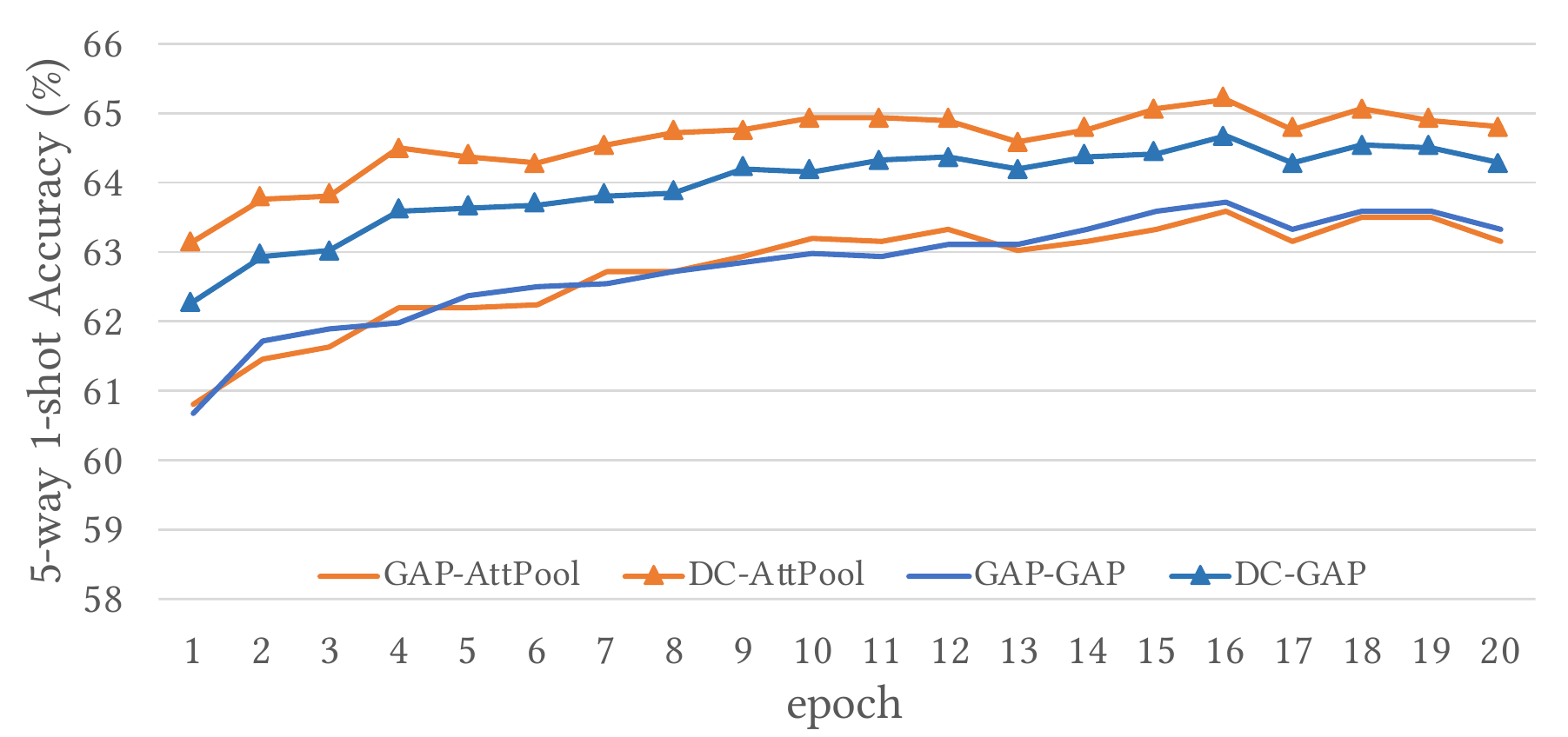}
} % resizebox end
\caption{\textbf{\textit{5-way 1-shot} accuracy on \textit{mini}Imagenet with meta-finetuning.} Dense classification pre-trained models are marked with $\triangle$. Models that utilize the same pooling in meta-finetuning are colored the same. Best viewed in color.}
\label{fig:metafinetuning}
\end{figure}

\noindent\textbf{Is meta-finetuning stage necessary?} 
Chen~\etal~\cite{chen2020new} suggested that competitive performance could be obtained on top of a pre-trained model without meta-finetuning. 
In this work, we have different findings. 
After the pre-training and right before meta-finetuning, we apply GAP over feature maps and use the resulting embeddings to perform few-shot classification as described in \S\ref{sec:protonetwork}.
The mean accuracies over 1000 meta-testing tasks are reported in Tab~\ref{tab:ablation_metafinetuning}.
It is astonishing to observe that the both DC-pretrained and GAP-pretrained models perform just slightly better than random guessing or even worse than random initialized models. 
% This is counterintuitive. If the pre-trained embeddings are not discriminative, how come the pre-training succeeds in the end? And conversely, if the pre-trained embeddings are discriminative, how come the extremely simple nearest-centroid classifier fails then? To figure out what is going on, one has to compare the cosine-similarity-based classifier (\eg Prototypical Networks) to the projection-based classifier used in our work. Maybe two images are close in one metric but far apart in the other. The finding is that the pre-trained model can achieve $>55\%$ accuracies if we turn off the \textit{track_running_stats} parameter of batch normalization layers. By disabling running mean and variance, the batch normalization layers use batch statistics in the meta-testing phase.
On the contrary, Fig~\ref{fig:metafinetuning} shows that the models can achieve $>60\%$ accuracies after meta-finetuning for one epoch. It indicates that meta-learning/meta-finetuning is at the core of the success of few-shot classification.

\begin{figure}[t]
% \vspace{-1em}
\resizebox{0.8\linewidth}{!}{ % resizebox begin
    \includegraphics[]{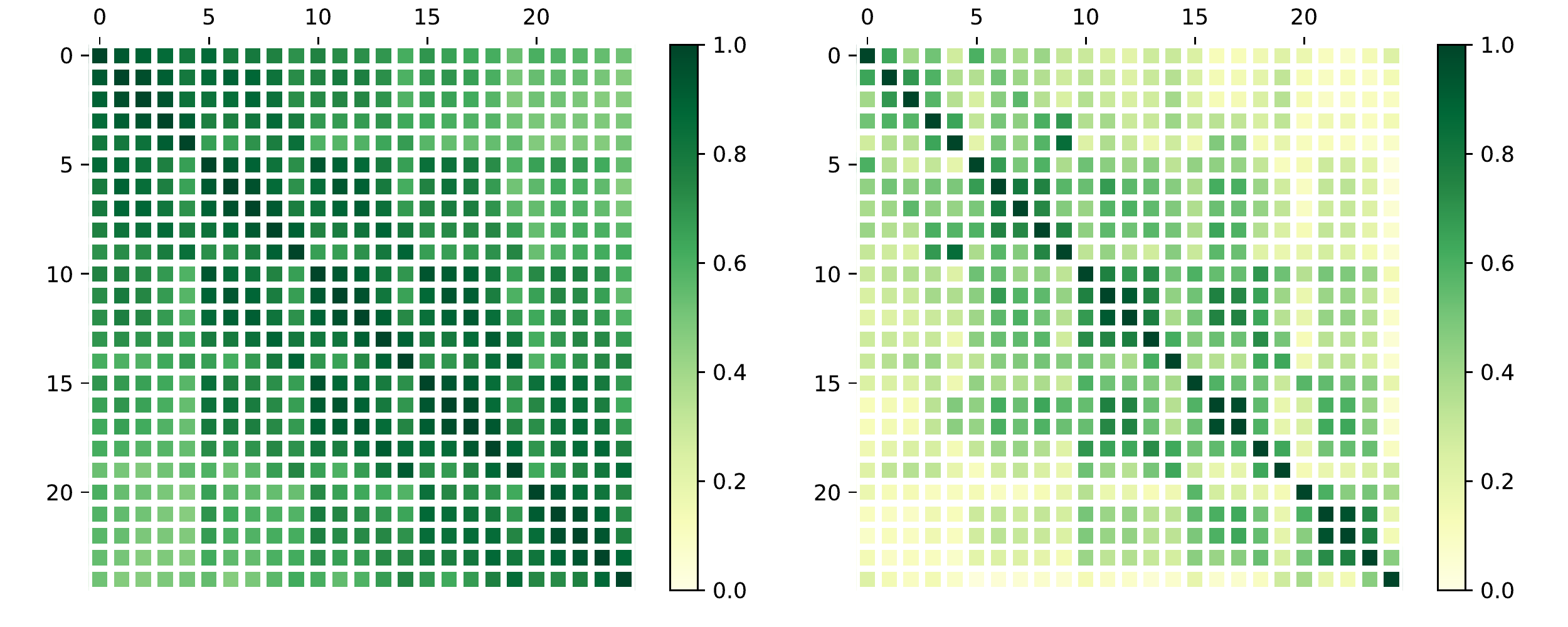}
} % resizebox end
% \vspace{-1.5em}
\caption{\textbf{Cosine similarities between local descriptors of a test image.} Left: similarities between local descriptors in the feature map generated by DC-pretrained model. Right: similarities between local descriptors in the feature map generated by GAP-pretrained model.}
\label{fig:sim_localactivations}
\end{figure}

\noindent\textbf{Influence of dense classification pre-training.} 
Tab~\ref{tab:ablation_twostagelearning} and Fig~\ref{fig:metafinetuning} show that meta-finetuning a DC-pretrained model generally achieves higher accuracies. 
To identify the reason, we investigate the effect of dense classification and find that the dense classification pre-training is able to produce smoother feature maps. 
The smoothness is evidenced by \textit{(i)} local descriptors in the feature map generated by the DC-pretrained model have $L2$-norm values with a smaller deviation, and \textit{(ii)}, as shown in Fig~\ref{fig:distractor} and Fig~\ref{fig:sim_localactivations}, a local descriptor in the feature map generated by the DC-pretrained model has higher cosine similarities with other local descriptors in its 8-connected neighborhood, demonstrating some kind of semantic consistency.
We hypothesize it is exact the semantic inconsistency between local descriptors in GAP-pretrained feature maps that makes it harder to meta-finetune for fast adaptation. 

\begin{table}[t]
% \caption{\textbf{\textit{5-way 1-shot} accuracy on \textit{mini}Imagenet with different loss weights.} We report the accuracies when meta-training from scratch with different loss weights. $\eta$ and $\gamma$ balance the effects of meta-learning and global classification.}
\caption{\textbf{5-way 1-shot accuracy on \textit{mini}Imagenet with different loss weights.} We report the accuracies when meta-training from scratch with different loss weights. $\eta$ and $\gamma$ balance the effects of meta-learning and global classification.}
\centering
\small
% \vspace{0em}
% \resizebox{0.88\linewidth}{!}{ % resizebox begin
\setlength{\tabcolsep}{5.5pt}
\begin{tabular}{ ccc|ccc}
\hline
\toprule
\textbf{$\mathcal{L}^{meta}(\eta)$} & \textbf{$\mathcal{L}^{ce}(\gamma)$} &&& \textbf{GAP} & \textbf{AttPool} \\
\midrule
    1.0 & 0.5 &&& 59.25 $\pm$ 0.70 & 56.69 $\pm$ 0.68 \\
    0.5 & 1.0 &&& 59.21 $\pm$ 0.65 & 58.42 $\pm$ 0.70 \\
    1.0 & 0.0 &&& 56.39 $\pm$ 0.68 & 56.51 $\pm$ 0.74 \\
\bottomrule
\end{tabular}
% } % resizebox end
\label{tab:ablation_loss_weight}
\end{table}

\begin{figure*}[t]
    % \vspace{-1em}
    \resizebox{\linewidth}{!}{
        \includegraphics[]{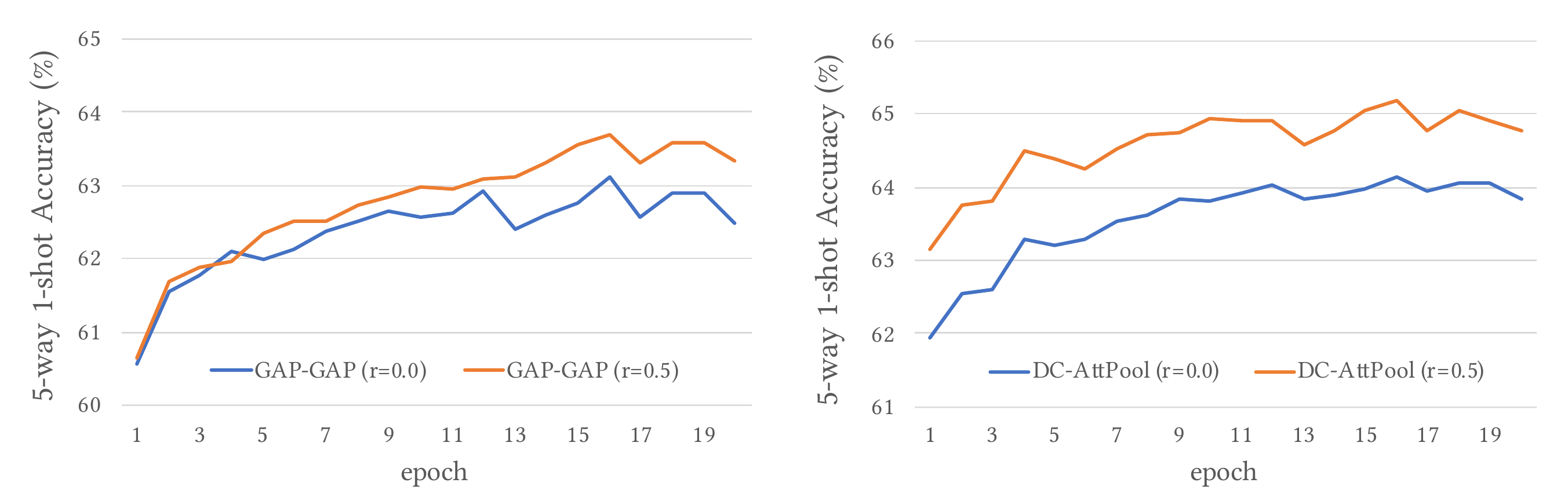}
    }
    \caption{\textbf{\textit{5-way 1-shot} accuracy on \textit{mini}Imagenet with/without global classification.} Left: meta-finetuning the GAP-GAP model with or without global classification. Right: meta-finetuning the DC-AttPool model with or without global classification. $\gamma$ is the loss weight of global classification in (\ref{eq:metafinetune}).}
    \label{fig:globalclassification}
\end{figure*}

\noindent\textbf{Influence of attentive pooling.} 
Attentive pooling aims to select informative local descriptors for constructing better embeddings for few-shot classification.
As can be seen in Tab~\ref{tab:ablation_twostagelearning}, applying attentive pooling when meta-finetuning a DC-pretrained model for few-shot classification gives the best \textit{5-way 1-shot} accuracy on \textit{mini}Imagenet. 
However, we notice that performance degradation happens when meta-training from scratch ($59.25\%$ vs $56.69\%$).
We conjecture that the reason is two-fold.
First, dense classification over local descriptors is much more difficult than standard image classification over image-level representations. 
As a result, Zero-AttPool struggles harder than Zero-GAP to balance the global classification and the few-shot classification when meta-trained from scratch, resulting in lower accuracy.
We confirm this finding by disabling global classification (\ie $\gamma = 0$) in meta-training. 
As shown in Tab~\ref{tab:ablation_loss_weight}, when the global classification is disabled, Zero-AttPool runs neck-and-neck with Zero-GAP.
We also find that the performance gap between Zero-AttPool and Zero-GAP can be reduced if increasing the loss weight $\gamma$ of global classification.
Second, the attention regressor cannot be efficiently optimized in meta-training from scratch.
This is because local descriptors in the very beginning are of no semantic information.
From the chaos, it is challenging for the regressor to figure out which descriptor is more important.
In other word, the regressor is error-prone which does harm to the global classification and the few-shot classification.
Only when starting from a pre-trained state, models can benefit from attentive pooling.
% We also notice that no improvement over GAP-GAP is observed when meta-finetuning a GAP-pretrained model ($63.58\%$ vs $63.57\%$). 

\noindent\textbf{Influence of global classification.}
We perform global classification in parallel with few-shot classification to enhance the representation ability of feature embeddings and encourage local descriptors to maintain their semantics during meta-training.
Tab~\ref{tab:ablation_loss_weight} shows that the joint learning is beneficial when meta-training from scratch.
To investigate the effect of global classification in our two-stage training, we examine models without global classification and plot accuracies in Fig~\ref{fig:globalclassification}.
As shown in the figure, global classification can consistently bring improvements when dealing with both the GAP-GAP model and the DC-AttPool model. 
Especially, the effect of global classification on our DC-AttPool is more obvious.

\subsection{Visualization}
\begin{figure}[t]
    \resizebox{0.92\linewidth}{!}{
        \includegraphics[]{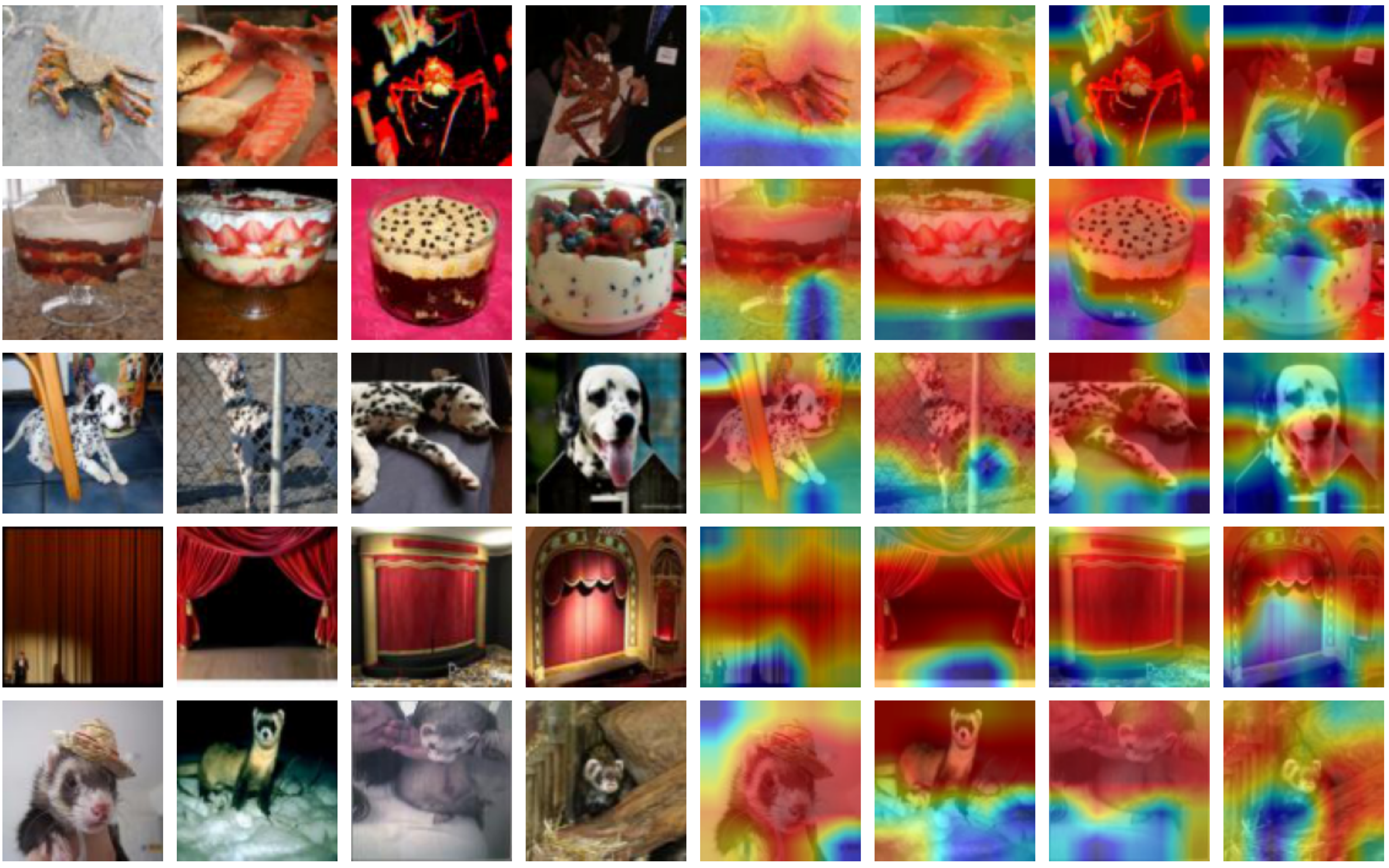}
    }
    \caption{\textbf{An example of the attention maps generated by the attention regressor.} The regressor can produce attention maps that focus on the target object in image and diminish the impact of background. In each row, we report a failure case in the last column. Warmer color indicates higher score.}
    \label{fig:attention_map}
\end{figure}

Our method highly relies on the attention regressor to correctly estimate the importance of each local descriptor within a feature map in order to select informative ones for few-shot classification. 
To check whether the regressor works as expected, we investigate the generated attention maps. 
Unsurprisingly, we find that the regressor can learn to concentrate on the target object in image by assigning higher attention scores to the corresponding descriptors.
And conversely, local descriptors that are related to distractors or the background normally get lower attention scores.
For visualization, we present an example of the attention maps generated by the regressor in Fig~\ref{fig:attention_map}.
As illustrated, for the three images from class \textit{crab}, the attention maps can capture the spatial information and highlight the target objects properly.
But we have to admit that the attention regressor is not perfect yet. 
It sometimes makes mistakes in which it fails to make an estimation that is consistent with our awareness.
We conjecture that the failure results from illumination, view-point, \etc. 

\section{Conclusion and Future Work}
We propose a new method, namely DCAP, to tackle the few-shot classification problem. 
Instead of designing complex classifiers as in prior art, in this work, we propose to adopt a simple nearest-centroid classifier and approach few-shot classification from the perspective of acquiring more discriminative and robust embeddings. 
Our DCAP is optimized following a two-stage training paradigm.
In the pre-training stage, we propose to train the embedding network to solve dense classification problem, and in meta-finetuning we fine-tune the pre-trained model on a family of few-shot tasks to perform few-shot classification.
An attention regressor is further devised in meta-finetuning to reweight local descriptors when summarizing them into image-level embeddings for few-shot classification.
We find that dense classification effectively eliminates semantic discrepancy among local descriptors that is observed in previous work. And the attention regressor learns to select informative local descriptors to build better embeddings.
By meta-finetuning a dense classification pre-trained model with attentive pooling, we reveal that the simple nearest-centroid classifier is able to yield accuracies that exceed most of the leading methods on two standard benchmark datasets. In the future, we will try to apply our proposed method for other visual tasks, such as few-shot visual dialog~\cite{guo2020iterative,guo2021context} and few-shot visual relationship detection~\cite{shang2019annotating,li2021interventional}. 

\section*{Acknowledgment}
This research was supported by the National Key R\&D Program of China under Grant 2019YFA0706203, the National Natural Science Foundation of China (NSFC) under grants 61932009 and 61976076, and A*STAR under its AME YIRG Grant (Project No. A20E6c0101). The authors would like to thank Shengen Tang, Huixia Ben, as well as anonymous reviewers for their careful reading of the manuscript and for providing constructive feedback and valuable suggestions.

\bibliographystyle{ACM-Reference-Format}
\bibliography{main}

\end{document}